\documentclass{article}

\usepackage{graphicx} % Required for inserting images

\usepackage{hyperref}
\usepackage{amsmath}
\usepackage{amssymb}
\usepackage{multirow}

\usepackage{tikz}
\usetikzlibrary{matrix}

\usepackage[utf8]{inputenc}
\usepackage{scrextend}

\usepackage{graphicx}
\usepackage{subcaption}

\usepackage{colortbl}
\usepackage{pifont}
\usepackage{array}
\usepackage{multirow}
\usepackage{xcolor}

\usepackage{booktabs}

\usepackage{algorithm}
\usepackage{algpseudocode}
\usepackage{footnote}

\usepackage{mdframed}

\usepackage{empheq}
\usepackage{fancybox}

\usepackage{hyperref}

\usepackage{pgfplots}
\pgfplotsset{compat=1.16} 

\usepackage{bm}

\usepackage[title]{appendix}
\usepackage{etoolbox}
\def\email#1{{\texttt{#1}}}
\providecommand{\keywords}[1]
{
  \small	
  \textbf{\textit{Keywords---}} #1
}

\title{Variational Inference Using Material Point Method}
\author{Yongchao Huang \footnote{Author email: \email{yongchao.huang@abdn.ac.uk}}}
\date{July 2024}

\begin{document}

\maketitle

\begin{abstract}
    A new gradient-based particle sampling method, \textit{MPM-ParVI}, based on material point method (MPM), is proposed for variational inference. MPM-ParVI simulates the deformation of a deformable body (e.g. a solid or fluid) under external effects driven by the target density; transient or steady configuration of the deformable body approximates the target density. The continuum material is modelled as an interacting particle system (IPS) using MPM, each particle carries full physical properties, interacts and evolves following conservation dynamics. This easy-to-implement ParVI method offers deterministic sampling and inference for a class of probabilistic models such as those encountered in Bayesian inference (e.g. intractable densities) and generative modelling (e.g. score-based). 
\end{abstract}

\keywords{Material point method; interacting particle system; sampling; variational inference.}

\section{Introduction}

Solid and fluid material simulations can be carried out using mesh-based or mesh-free approaches. Mesh-based methods, e.g.  finite element method (FEM \cite{Courant1943}), finite difference method (FDM \cite{Courant1928}), and finite volume method (FVM \cite{Varga1962}), involve discretizing the space into a finite number of non-overlapping elements (i.e. a rigid mesh), which are popular but may not be applicable to large deformations as elements are distorted and accuracy lost \cite{DeVaucorbeil2020}. The mesh granularity trades off efficiency and accuracy, as discretized differential equations (e.g. the Navier-Stokes equations) are evaluated at grids. Computation suffers from curse of dimentionality as the number of particles increases exponentially in response to dimension increase. Meshless or mesh-free methods \cite{Liu2002}, e.g. smoothed particle hydrodynamics (SPH \cite{Gingold1977,Lucy1977numerical}) and material point method (MPM \cite{Sulsky1994,Sulsky1996}), avoid these issues by discretizing the space into a set of particles \footnote{Discretizating a material into a set of points is faster and less complex than the meshing process in mesh-based methods \cite{DeVaucorbeil2020}.}, and each particle interacts with its neighboring particles in a flexible manner \cite{DeVaucorbeil2020}. MPM represents one of the latest developments in particle-in-cell (PIC \cite{Harlow1964,Brackbill1986,Zhu2005}) methods developed in the early 1950s for simulating solids \cite{Sulsky1995,Li2000,Lee2010,Lian2011b,Iaconeta2019,Leavy2019}, fluids \cite{Brackbill1986,York1997,York2000,Gilmanov2008b,Lian2014,Kularathna2017}, and gases \cite{Ma2009,Tran2010} and many other continuum materials. Recent years have witnessed vast developments and applications of MPM and its variants in physics \cite{Lu2006,Jiang2015thesis}, engineering \cite{York2000,Nairn2003,Ionescu2005,Beuth2007,Daphalapurkar2007,LlanoSerna2016,Fern2019} and computer graphics \cite{Zhu2005,Stomakhin2013,Jiang2015thesis,Jiang2016,Jiang2017} for simulating physical behaviours of deformable materials such as elastic, plastic, elasto-plastic and viscoelastic materials, snow, lava and sand, etc. 

Representing a continuum as a set of discrete particles and evolving (i.e. deforming or flowing) them according to exact or approximate physics \footnote{Typically continuity equations that specify conservation of mass, momentum, and energy.} motivates the design of new particle-based variational inference (ParVI) algorithms. Among these are the recently proposed physics-based ParVI algorithms which formulate an inference problem as a deterministic physical simulation process, following principles of electrostatics (e.g. \textit{EParVI} \cite{huang2024EParVI}) or fluid mechanics (e.g. \textit{SPH-ParVI} \cite{huang2024SPHParVI}). Spotting the expressive power of MPM in representing and calculating an interacting particle system (IPS), we can, with a similar spirit of SPH-ParVI, modify MPM for use of inferring probability densities. This can be achieved by incorporating the target density as a virtual, external force field applied to the particles in MPM simulation. MPM evolves the particles towards a transient or steady state which resembles the target geometry following embedded physics. Simulation correctness \footnote{Existence, convergence and uniqueness of such an optimal state which approaches the target shape demands rigorous mathematical proof.} and tractability are ensured by the well-established MPM mechanism, although assumptions and approximations may be set to trade off accuracy and efficiency.

Statistical inference is relevant to many machine learning tasks. In diffusion and generative modelling, for example, estimating the gradients of the log density is a prerequisite. Density \footnote{By density, we refer to the \textit{probability density function} (pdf).} estimation plays a central role in modern probabilistic approaches. Bayesian inference, for example, yields uncertainty quantification which is beneficial for many decision-making scenarios (e.g. probabilistic state estimation in reinforcement learning). In general, densities can be obtained either from an exact, analytic probabilistic model (e.g. a Bayesian posterior), or, in the presence of analytical intractability (e.g. high dimensional integration), via approximate, numerical schemes such as Markov chain Monte Carlo (MCMC) sampling or variational inference. The later have been successful and versatile though, still, efficient and accurate inference methods are in demand, particularly for inferring complex (e.g. multi-modal), high-dimensional geometries. 

\section{Related work}

As this work concerns about MPM and inference methods, we introduce some related work on both aspects. For clarity, readers can refer to a full list of the abbreviations of terminologies used in this context in Appendix.\ref{tab:terminologies}.

\paragraph{Meshless particle simulation methods} Meshless particle methods divide a material into a set of discrete particles. They are well suited for simulations of granular materials with complex geometries because body-fitted meshes are not needed \cite{DeVaucorbeil2020}. Some \footnote{For a comprehensive review, see e.g. \cite{DeVaucorbeil2020, Zhang2016book}.} of the meshless methods include SPH \cite{Gingold1977,Lucy1977numerical}, the generalized finite difference method \cite{Liszka1980}, the discrete element method (DEM \cite{Cundall1979DEM}, the diffuse element method (DEM \cite{Nayroles1992}), the element free Galerkin method (EFG \cite{Belytschko1994}), the material point method (MPM \cite{Sulsky1994}), the reproducing kernel particle method (RKPM \cite{Liu1995}), the natural element method \cite{Sukumar1998}, the meshless local Petrov Galerkin method (MLPG \cite{Atluri1998}), the particle finite element method (PFEM \cite{Idelsohn2006}), the optimal transport meshfree method (OTM \cite{Li2010}), etc. Meshfree methods in general avoids element distortion and expensive remeshing during material deformation.

Particle-in-cell (PIC) methods were developed by Harlow \cite{Harlow1964}, primarily for applications in fluid mechanics \cite{DeVaucorbeil2020}. Later Brackbill and Ruppel \cite{Brackbill1986,Brackbill1988} introduced the fluid implicit particle (FLIP) method which improves early PIC methods. The FLIP method was further developed by Sulsky et al. \cite{Sulsky1994,Sulsky1995} for applications in solid mechanics which has since been termed the material point method (MPM). While PIC and MPM use both Lagrangian particles and a background Eulerian grid \footnote{The background lattice can be a regular Cartesian or unstructured grid which the simulated body occupies during deformation or flow. There are MPM variants, e.g. TLMPM \cite{deVaucorbeil2019}, uses background grid that only cover a reference space.} simultaneously, particles in PIC carry only position and mass, MPM allows particles to carry full physical state \cite{DeVaucorbeil2020}. Some MPM variants include the generalized interpolation material point method (GIMP \cite{Bardenhagen2004,Ma2006}), B-splines MPM (BSMPM \cite{Steffen2008}), the convected particle domain interpolation (CPDI \cite{Sadeghirad2011}), the dual domain MPM (DDMPM \cite{Zhang2011}), weighted least square MPM \cite{Wallstedt2011}, moving least square MPM \cite{Edwards2012}, improved MPMs (iMPM \cite{Sulsky2016,Wobbes2019,Tran2020}), the total Lagrangian MPM (TLMPM \cite{deVaucorbeil2019}) and discontinuous Galerkin MPM (DGMPM \cite{Renaud2018,Renaud2020}, isogeometric MPM (IGA-MPM \cite{Moutsanidis2020}), local maximum-entropy MPM (LME-MPM \cite{Perez2021}), affine matrix-based MPM (AM-MPM, \cite{He2023}), etc. To model fluids and gases, weakly compressible \cite{York1999} and incompressible \cite{Stomakhin2014} MPMs have also been developed. These variants differ in the basis functions used (e.g. $C^0$ or $C^1$ smoothness), the time integration scheme used (explicit or implicit solvers \footnote{Explicit MPM algorithm is simpler and easier to implement, while implicit integration schemes are more expensive due to construction of the tangent stiffness matrix \cite{DeVaucorbeil2020}.}), the type of grid used (uniform Cartesian grid or unstructured mesh \footnote{If an unstructured grid is used, neighbourhood search needs to be performed during particle-mesh interaction, which is expensive. However, it facilitates enforcement of boundary conditions \cite{DeVaucorbeil2020}.}), and so on.

SPH, developed in the 1970s and 1980s \cite{Gingold1977}, is a fully meshfree method; it is applicable to a space with arbitrary dimensions. Particles in SPH directly interact with each other (through the smoothing kernel), while in MPM, particle interactions are propagated through the background Eulerian grid \footnote{The equations of motion, i.e. balance of momentum, are solved at grid nodes \cite{Steffen2008}.}. Both SPH and MPM are used in continuum mechanics to model solids and fluids at macroscopic level, and variables can be calculated using smooth functions over the space \cite{DeVaucorbeil2020}. It has been shown \cite{Ma2009b,DeVaucorbeil2020} that MPM and SPH predictions are quite similar, but MPM is faster \footnote{The size of time step used in MPM is much larger than that of SPH. Also, no neighbourhood particle search is used in MPM. Directly computing particle interactions takes $\mathcal{O}(M^2)$ \cite{huang2024SPHParVI}.} (as it avoids direct computation of particle interactions) and more accurate than SPH. Also, MPM simulation requires less tricky ad hoc parameters than SPH \cite{DeVaucorbeil2020}; typical hyper-parameters for MPM are the mesh spacing and time step. A combined SPH-MPM method, which exhibits superior performance over SPH and MPM, has been developed \cite{Raymond2018,He2019}. There are efforts \cite{Zhang2016book} to develop hybrid methods which couple MPM with other particle modelling methods for improved efficiency, e.g. the explicit material point finite element method (MPFEM \cite{Zhang2006}), molecular dynamics with MPM \cite{Lu2006}, and so on. 

\paragraph{ParVI methods} Particle-based variational inference (ParVI) methods provide some of the most flexible and expressive density inference methods. They don't pre-assume any parametric form for the variational distribution, instead, they issue a set of particles and evolve them towards the target configuration following dynamics driven by the target. Stein variational gradient descent (SVGD \cite{Liu2016SVGD}), for example, evolves a set of particles from an initial configuration (the prior distribution) till a converged state, following the gradient field of the log target density \footnote{In SVGD, the gradient term serves as the driving (dominating) force, while the kernel differential term acts as a repulsive force. Both are essential to capture the target, maintain particle diversity and to avoid mode collapse.}. ParVI methods are capable of capturing complex geometries such as those with high curvature, heavy-tail and multi-modal characteristics. Most, if not all, of them can scale to large datasets and high dimensions - they in general don't suffer much from curse of dimensionality.

Conventional ParVI methods, e.g. SVGD \cite{Liu2016SVGD}, sequential Monte Carlo (SMC \cite{Doucet2001}), particle-based energetic variational inference (EVI \cite{Wang2021EVI}), etc, explicitly or implicitly optimize certain discrepancy measure (e.g. KL divergence or ELBO \cite{VI_Blei}) based on statistical principles (e.g. the Stein's identity). Purely physical simulation based ParVI approaches are recently proposed. Huang \cite{huang2024EParVI} designed the EParVI algorithm which applies electrostatics principles to evolve a system of charges under a target electric field. This method is tested effective on low dimensional inference problems including Bayesian logistic regression and dynamic system identification. Huang \cite{huang2024SPHParVI} proposed the SPH-ParVI method which uses the SPH method to simulate particle movements, under external pressure or force field aligned with the target. These recently invented physics-based ParVI methods formulate the sampling problem as a physical simulation process which generates (quasi) samples at transient or terminal time. The simulation process is governed by physical laws such as conservation of mass, momentum and energy. Mass conservation is, in general, automatically satisfied in these particle-based simulations, as each particle carries constant amount of mass over time. Particle movement is described by, depending on the physics adapted, e.g. the Poisson's equation \cite{huang2024EParVI}, Navier-Stokes equations \cite{huang2024SPHParVI}, and so on.

MPM, as a particle simulation methodology, has not been applied in any probabilistic machine learning setting. We therefore extend MPM to arbitrary dimensions and devise the MPM-ParVI method for variational inference. This work is by no means an innovation of the MPM method, but rather an application to the area of statistical inference. In the following, we describe the fundamental principles of MPM and demonstrate how it can be applied to statistical sampling and inference; experiments to validate the proposed methodology will be presented in future work.

\section{MPM-based sampling: methodology}

The basic problem setting is, we have a deformable body, e.g. a solid or fluid, characterized by its initial position $\mathbf{X}$, current position $\mathbf{x}$, displacement $\mathbf{u}(\mathbf{X},t)$, velocity $\mathbf{v}(\mathbf{X},t)$, acceleration $\mathbf{a}(\mathbf{X},t)$, as well as other physical quantities derived from these. Under the action of an external force $\mathbf{f}^{ext}(\mathbf{X})$ field \footnote{One can also introduce an external pressure field, e.g. \cite{huang2024SPHParVI}, or initial/boundary conditions to the simulation.}, we would like to know the configuration of the body, i.e. distribution of particles in the language of MPM, at a transient or terminal time $t$. This transient or steady state of the deformable body can be numerically predicted following deterministic physics such as continuum mechanics (solid or fluid mechanics in particular). The key question thus lies in the efficiency and accuracy of the physical simulation, as modelling the motion of an \textit{interacting particle system} (IPS) involves solving (partial) differential equations. Particle-based simulation methods such as SPH and MPM are some of the most computationally efficient numerical techniques for this kind of simulation with balanced accuracy, against the long standing mesh-based methods such as FEM, FVM, FDM and many others.

Observe that ParVI methods also model an IPS, i.e. evolving a set of particles towards a target distribution. We can therefore introduce the target density $p(\mathbf{x})$ as an external, driving effect, and formulate the probabilistic inference problem as a deformation simulation procedure. The key lies in how to effectively and efficiently evolve the interacting particle system (IPS) to form the target geometry. Using established theory of mechanics to evolve an IPS has the advantages of deterministic physics, flexibility of formulation (i.e. pressure, force, boundary conditions) and performance (e.g. efficiency, convergence) guarantee. In the following, we first present the basic MPM principles, then modify it for probabilistic inference. As MPM is originated from continuum mechanics, interested readers are referred to prerequisites such as kinematics, kinetics, constitutive models (e.g. elasticity), and conservation laws. Here we briefly present the results, more details can be found in Appendix.\ref{app:MPM_more_details}, as well as good references such as \cite{Jiang2015thesis,Jiang2016,DeVaucorbeil2020}.

\paragraph{MPM formulation} 

\begin{figure} [H]
    \centering
    \includegraphics[width=0.45\linewidth]{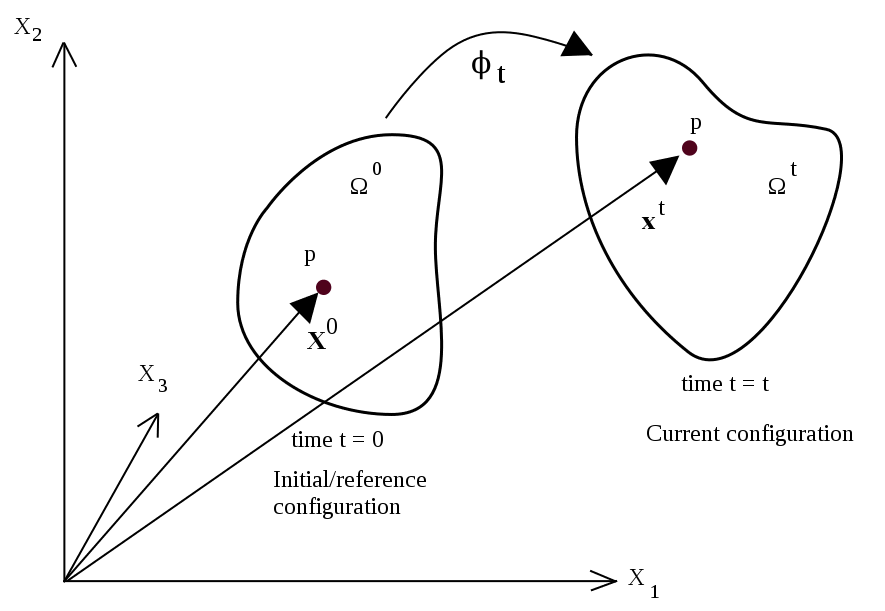}
    \caption{MPM description of motion of a continuum body in 3D space (figure from \cite{Kumar2022}). $\Omega^0$ and $\Omega^t$ are the initial and current configurations of the whole material domain, respectively. $p$ represents a material point (i.e. a particle). $\mathbf{X}^0$ and $\mathbf{x}^t$ are the initial and current positions of the material point, respectively. $\phi_t$ is the mapping function between initial and current configurations.}
    \label{fig:particle_motion}
\end{figure}

MPM is a \textit{hybrid Eulerian-Lagrangian} method for modelling material deformation, with particles (arguably) being the primary material representation \cite{Jiang2015thesis}. Two coordinate systems, i.e. the \textit{Lagrangian} or \textit{material} coordinate $\mathbf{X}$ and the \textit{Eulerian} or \textit{spatial} coordinate $\mathbf{x}$, are employed in MPM to coherently describe particle motions and material deformation \footnote{The Lagrangian coordinate $\mathbf{X}$ are fixed to the material points \footnote{Material points and particles are interchangeably used; nodes, grid points, mesh points are also interchangeably used in MPM.} and follows individual material points as they move through space and time. Particle positions are described relative to their initial configuration; motion of each particle is tracked explicitly by updating its coordinates as it moves. The Eulerian coordinates are fixed in space and material flows through the spatial grid.}. More details about Lagrangian and Eulerian methods can be found in Section.\ref{sec:discussions}. In MPM, each Lagrangian particle \footnote{To be consistent with notations used in other MPM literature, we use the subscript $p$ to represent particle and denote its index, $p=1,2,...,M$, where $M$ is the total number of particles. We use the subscript $i$ to index grid node, with $i=1,2,...,N$, where $N$ is the total number of mesh nodes. Some literature such as \cite{Jiang2016} use bold $\mathbf{i}$ to represent multi-index. In this context, we use plain $i$ to represent node identity, and use a regular Cartesian coordinate system for the grid nodes $\mathbf{x}_i$.} $p$ carries physical properties such as position $\textbf{x}_p$, mass $m_p$, density $\rho_p$, velocity $\textbf{v}_p$, deformation gradient $\textbf{F}_p$, Cauchy stress tensor $\bm{\sigma}_p$, and other state variables \cite{DeVaucorbeil2020}. A background Eulerian grid, with certain meshing space $h$, is defined to discretize the continuous fields. The Lagrangian coordinate is used to track particle positions and states, while the Eulerian grid is used to solve the governing equations. Material points interact with the grid, transfer their states of properties to grid nodes for computation and then update their states based on grid solutions. 

The motion of a deformable body is schematically shown in Fig.\ref{fig:particle_motion}. Initially particles are at position $\mathbf{X}$ at time $t$=0, at current time $t$ they are located at position $\mathbf{x}$. In Lagrangian description, these two configurations are related through a mapping function $\phi$:

\begin{equation} \label{eq:cc_Lagrangian_Eulerian_coordinates} \tag{cc.Eq.\ref{eq:Lagrangian_Eulerian_coordinates}}
    \mathbf{x} = \phi (\mathbf{X},t)
\end{equation}

\noindent and we have $\mathbf{X}=\phi(\mathbf{X},0)$. Common physical quantities used to describe motion, e.g. \textit{displacement} $\mathbf{u}$, \textit{velocity} $\mathbf{v}$, \textit{acceleration} $\mathbf{a}$ and the \textit{deformation gradient} $\mathbf{F}$, can be expressed as: 

\begin{equation} \label{eq:displacement_velocity_acceleration_deformationGradient} \tag{cc.Eq.\ref{eq:displacement} $\sim$ \ref{eq:deformation_gradient}}
\begin{aligned}
    & \mathbf{u}(\mathbf{X},t) := \phi(\mathbf{X},t) - \phi(\mathbf{X},0) = \mathbf{x} - \mathbf{X}, \quad \mathbf{v}(\mathbf{X},t) := \frac{\partial \phi (\mathbf{X},t)}{\partial t} \\
    & \mathbf{a}(\mathbf{X},t) := \frac{\partial \mathbf{v} (\mathbf{X},t)}{\partial t} = \frac{\partial^2 \phi (\mathbf{X},t)}{\partial t^2}, \quad \mathbf{F}(\mathbf{X},t) := \frac{\partial \phi(\mathbf{X},t)}{\partial \mathbf{X}} = \frac{\partial \mathbf{x}(\mathbf{X},t)}{\partial \mathbf{X}} \\ 
\end{aligned}
\end{equation}

Particle movements, or body deformations, are governed by conservation and constitutive equations. Two \footnote{A third conversation equation, i.e. the energy equation, concerns about thermo-mechanical behaviours, which is not relevant in this context.} governing equations of interest are the conservation of mass and (linear) momentum:

\begin{equation} \label{eq:conservation_of_mass_and_momentum} \tag{cc.Eq.\ref{eq:conservation_of_mass}, \ref{eq:conservation_of_linear_momentum}}
\begin{aligned}
    & \frac{D \rho}{D t} + \rho \nabla \cdot \mathbf{v} = 0 \quad \text{(Continuity equation)} \\
    & \rho \frac{D \mathbf{v}}{D t} = \nabla \cdot \bm{\sigma} + \mathbf{f}^{ext} \quad \text{(Linear momentum equation)} \\
\end{aligned}
\end{equation}

\noindent where $\rho(\mathbf{X},t)$ is the density, $\bm{\sigma}(\mathbf{X},t)$ is the symmetric Cauchy stress tensor, $\mathbf{f}^{ext}$ is the external body force \footnote{Body forces, such as gravity, are forces that act on the entire volume of the material. Body forces are usually defined per unit mass or per unit volume.} per unit volume, e.g. the gravitational force $\mathbf{f}^{ext}=\rho \mathbf{g}$ with $\mathbf{g}$ being the gravitational acceleration. $\frac{D \rho}{Dt} = \frac{\partial  \rho}{\partial t} + (\nabla  \rho) \cdot \mathbf{v}$ is the \textit{material derivative} \footnote{Also termed \textit{substantial derivative} or \textit{particle derivative}.} of the density. Conservation of mass is also called \textit{mass continuity}, which is automatically satisfied for an IPS as particles carry constant mass. The momentum equation, which echoes the \textit{Newton's second law} on a continuum body, governs the motion of particles. 

Given the governing equations, we are now faced with two questions for deforming a material: first, how to derive the quantities, e.g. stress $\bm{\sigma}$, from an observable quantity such as particle position $\mathbf{X}$ or associated quantities? Second, how to solve the differential equations in discrete time and space? The first question can be solved by a \textit{constitutive model} which relates stress to a deformation field. The second concerns about the discretization of the differential equations. There exist explicit and implicit time integration schemes to compute these quantities on a grid of mesh points at discrete times. MPM, in particular, calculates some of these quantities on a grid and propagates them back to the particles.

MPM relies on a proper constitutive model to relate stress and strain, or force and deformation. A constitutive relation is similar to the equation of state (EoS) used in SPH, which defines how a material responds to internal and external effects. It differentiates different types of materials, e.g. elastic, hyper-elastic, plastic, visco-elastic, elasto-plastic, etc. In particular, it distinguishes solids and fluids behaviours. The constitutive model for \textit{isotropic}, \textit{linear elastic materials}, for example, writes:

\begin{equation} \label{eq:cc_stress_strain_linear_elastic_material} \tag{cc.Eq.\ref{eq:stress_strain_linear_elastic_material}}
    \boldsymbol\sigma = (\lambda \text{tr} \boldsymbol\epsilon) \mathbf{I} + 2\mu \boldsymbol\epsilon
\end{equation}

Constitutive relation can be derived from the \textit{energy density function} $\Psi$ which is normally defined as a function of the deformation gradient $\mathbf{F}$ in continuum mechanics. Hyper-elastic materials, for example, have

\begin{equation} \label{eq:cc_1stPK_stress_deformation_gradient} \tag{cc.Eq.\ref{eq:1stPK_stress_deformation_gradient}, \ref{eq:Cauchy_stress_deformation_gradient}}
    \mathbf{P} = \frac{\partial \Psi(\mathbf{F})}{\partial \mathbf{F}}, \quad 
    \boldsymbol{\sigma} = \frac{1}{\det(\mathbf{F})} \frac{\partial \Psi(\mathbf{F})}{\partial \mathbf{F}} \mathbf{F}^\top
\end{equation}

\noindent where $\mathbf{P}$ is the \textit{first Piola-Kirchoff stress} (1st PK), $\boldsymbol{\sigma}$ the \textit{Cauchy stress}, and $J=\det(\mathbf{F})$ is the determinant of the matrix (second-order tensor) $\mathbf{F}$. We observe $\boldsymbol{\sigma}=\mathbf{P} \mathbf{F}^{\top} / J$. Conversion between different types of stresses is presented in Table.\ref{table:stress_relations} of Appendix.\ref{app:MPM_more_details}.

For example, the energy density function of the \textit{Neo-Hookean} model, a simple non-linear hyper-elastic model for describing isotropic elastic materials (particularly for large deformations), is

\begin{equation} \label{eq:cc_Neo_Hookean_energy_density_func} \tag{cc.Eq.\ref{eq:Neo_Hookean_energy_density_func}}
\Psi(\mathbf{F}) = \frac{\mu}{2} \left( \text{tr}(\mathbf{F}^\top \mathbf{F}) - d \right) - \mu \log(J) + \frac{\lambda}{2} \log^2(J)
\end{equation}

\noindent where $d$ is the dimension. The first Lamé parameter (i.e. the shear modulus) $\mu$ and the second Lamé parameter $\lambda$ can be derived from the Young’s modulus and Poisson’s ratio of the material (see Eq.\ref{eq:shear_modulus_Lame_constant} in Appendix.\ref{app:MPM_more_details}). Differentiating $\Psi(\mathbf{F})$, we obtain the 1st PK stress, and then Cauchy stress, for Neo-Hookean materials \cite{Jiang2016}:

\begin{equation} \label{eq:cc_1stPKStress_CauchyStress_F} \tag{cc.Eq.\ref{eq:1stPKStress_CauchyStress_F}}
    \mathbf{P} = \mu (\mathbf{F} - \mathbf{F}^{-\top}) + \lambda \log(J) \mathbf{F}^{-\top}, \quad \boldsymbol{\sigma} = \frac{1}{J} [\mu (\mathbf{F}\mathbf{F}^\top - \mathbf{I}) + \lambda \log(J) \mathbf{I}]
\end{equation}

\noindent this internal \footnote{The internal force are induced by deformation gradients and in fact calculated at particles. They are propagated to grid nodes for solving the equation of motion at node points.} stress (denoted as $\mathbf{f}^{int}$), together with external force, is plugged in the momentum conservation equation (\ref{eq:conservation_of_mass_and_momentum}) to update particle velocities.

\paragraph{The interpolation function} Unlike in SPH and electrostatics which model the IPS via direct interactions, MPM models indirect interactions between particles via use of a background grid. One can imagine the Eulerian grid as a medium to propagate interactions; in fact, the interpolation function, as described here, plays a similar role as the kernel function used in SPH - they account for the influences of, and weight the contributions from, neighbourhood particles. 

At each iteration \footnote{Here and later, we use the term \textit{iteration} to represent a time step $t$, e.g. $t=n$ implies iteration $n$.}, MPM transfers the mass $m_p$ and momentum $m_p \mathbf{v}_p$ of all particles to grid nodes, and solves the momentum equation in \ref{eq:conservation_of_mass_and_momentum} to update nodal velocities $\mathbf{v}_i$, and maps them back to particles to update particle velocities $\mathbf{v}_p$, positions $\mathbf{x}_p$, etc. The \textit{interpolation functions}, also called \textit{interpolation kernels}, are used to transfer the states of properties between particles and grid nodes. An interpolation function $K(\cdot)$ accepts the relative distance $\mathbf{x}_i-\mathbf{x}_p$ between a grid node $i$ and a particle $p$ as input, and outputs the weight $w_{ip}=K(\mathbf{x}_i-\mathbf{x}_p)=K_i(\mathbf{x}_p)$ which determines how strongly the particle and node interact \cite{Jiang2016} - its value gets larger as they come closer. The weight for a 3D node $\mathbf{x}_i=(x_i,y_i,z_i)$ and particle $\mathbf{x}_p=(x_p,y_p,z_p)$, for example, can be evaluated using the dyadic products \cite{Steffen2008,Jiang2016} of a 1D kernel $K(\cdot)$:

\begin{equation} \label{eq:interpolation_weights}
    w_{ip}=K (\frac{1}{h}(x_p - x_i)) K(\frac{1}{h}(y_p - y_i)) K(\frac{1}{h}(z_p - z_i))
\end{equation}
where $h$ is the grid spacing. 

Correspondingly, the weight gradient $\nabla w_{ip} =\nabla K(\mathbf{x}_i - \mathbf{x}_p)=\nabla K_i(\mathbf{x}_p)$ is computed as:

\begin{equation} \label{eq:interpolation_weight_gradients}
\nabla w_{ip} = 
    \begin{pmatrix}
    \frac{1}{h} K'\left(\frac{1}{h}(x_p - x_i)\right) K\left(\frac{1}{h}(y_p - y_i)\right) K\left(\frac{1}{h}(z_p - z_i)\right) \\
    K\left(\frac{1}{h}(x_p - x_i)\right) \frac{1}{h} K'\left(\frac{1}{h}(y_p - y_i)\right) K\left(\frac{1}{h}(z_p - z_i)\right) \\
    K\left(\frac{1}{h}(x_p - x_i)\right) K\left(\frac{1}{h}(y_p - y_i)\right) \frac{1}{h} K'\left(\frac{1}{h}(z_p - z_i)\right)
    \end{pmatrix}
\end{equation}
where $K'(\cdot)=\nabla K(\cdot)$ is the derivative of the 1D kernel $K(\cdot)$. The interpolation weights \(w_{ip}\) and weight gradients \(\nabla w_{ip}\) are stored on particles. The interpolation weight \(w_{ip}\) is a scalar, representing the contribution of grid node $i$ to particle $p$; the weight gradient \(\nabla w_{ip}\) is a vector, informing how interpolation weight changes over space.

Choosing a proper interpolation kernel $K(\cdot)$ is crucial in MPM simulation. General considerations \cite{Jiang2016} include smoothness, numerical stability, computational efficiency (speed and memory) and width of the stencil. MPM typically requires the kernel to have $C^1$ continuity to avoid \textit{cell-crossing instability}. Using cubic splines can improve the spatial convergence of MPM as a result of reduced grid-crossing errors \cite{Steffen2008,Steffen2008c,DeVaucorbeil2020}. Cubic B splines are frequently used in animation simulations \cite{Stomakhin2013,Jiang2016}. Some kernels, e.g. the multi-linear kernel, may produce discontinuous gradients $\nabla w_{ip}$ and therefore discontinuous forces \cite{Jiang2016}. Three commonly used kernels are listed in Appendix.\ref{app:interpolation_kernels}.

\paragraph{Representing the target score as external body force} We now know how to calculate and transfer the quantities needed for solving the momentum equation (\ref{eq:conservation_of_mass_and_momentum}). The next step is to introduce external effects, i.e. information about the target density $p(\mathbf{x})$, to the dynamics, as our final goal is to move the particles to form the target geometry. In MPM, external forces can be conveniently applied on grid nodes and/or particles. For example, the gravitational force $\mathbf{f}_p^{ext} = m_p \mathbf{g}$, with particle mass $m_p$ and the gravitational acceleration $\mathbf{g}$, is normally applied in MPM simulations. To infer a target density $p(\mathbf{x})$ with $\mathbf{x} \in \mathbb{R}^d$, we can represent the gradient field of $\log p(\mathbf{x})$, also termed \textit{score} \cite{SM_Yang,SBGM_Huang} in statistics, as a time-invariant, external force field \footnote{To yield more realistic simulation, one can freely add the gravitational force, i.e. $\mathbf{f}^{ext} = m \mathbf{g} + \alpha \log p(\mathbf{x})$ to a grid node or particle.}, i.e. $\mathbf{f}^{ext}(\mathbf{x})=\nabla_{\mathbf{x}} \log p(\mathbf{x})$, and use MPM to evolve the IPS under internal and external effects \footnote{While external force is the driving force, internal forces such as inertial, viscous friction, etc, serve as repusive forces and are essential for modelling interactions and maintaining diversity.}. It is expected that, some transient or terminal (steady) configuration of the particles, i.e. $\Omega^t$ in Fig.\ref{fig:particle_motion}, shall approximate the target geometry, i.e. particles cluster and concentrate around high density regime(s) while maintaining certain degree of dispersion (diversity), as a result of external (dominating) and internal (e.g. friction or inertia, as determined by the constitutive model) effects. Formulating the inference problem as a physical simulation process guarantees some nice properties aligned with statistical inference: at an optimal state, particles shall distribute in space proportionally to the strength of the applied external driving force, and our design ensures particles flow towards high density regime(s) following the score field. Particles can distribute around multiple modes if the material constitutive model is properly defined (e.g. allowing for fracture). Particles won't collapse into single points (i.e. point estimates) unless the solid or fluid material being simulated is extremely deformable (e.g. negative Poisson's ratio). These guarantees, i.e. multiple-modality and high density concentration with diversity, are desired for designing an inference method for use in e.g. Bayesian inference and generative modelling where both accuracy and uncertainty are desired. In extra, employing physics to model the IPS yields deterministic inference. In fact, the physics-guided evolution process is implicitly minimizing the potential \footnote{This needs formal analysis in future work.} of the particle system - an objective similar to the KL divergence or ELBO as used in classic VI formulations.

For example, if we apply the following time-invariant, external body \footnote{In mechanics, body force distributes and acts across the entire volume of a body, while surface force acts on the surface of a body. Common body forces include gravitational force, electromagnetic force, and inertial (fictitious) force.} force on all grid nodes

\begin{equation} \label{eq:target_score_as_external_force}
    \mathbf{f}_i^{ext}(\mathbf{x}_i)=\alpha \nabla_{\mathbf{x}} \log p(\mathbf{x}_i),
\end{equation}

\noindent where $\alpha$, either positive or negative, is a magnitude amplification constant to weight the external force, in each iteration $t=n$, the combined external and internal forces can then be used to update grid velocities $\mathbf{v}_i$ via e.g. an explicit Euler time integration scheme: 

\begin{equation} \label{eq:Eulerian_time_integration}
    \mathbf{v}_i^{n+1} = \mathbf{v}_i^n + \Delta t \cdot \frac{(\mathbf{f}_i^{\text{int}} + \mathbf{f}_i^{ext})}{m_i}
\end{equation}
where $m_i$ is the nodal mass, $\Delta t$ is the time step size, and $\mathbf{f}_i^{\text{int}}$ is the propagated internal force induced by deformation gradients (\ref{eq:cc_1stPK_stress_deformation_gradient}). Using interpolation functions, the updated grid node velocities are then mapped back to particles (i.e. grid to particle, \textit{G2P}) to update particle velocities $\mathbf{v}_p$:

\begin{equation} \label{eq:cc_particle_velocity} \tag{cc.Eq.\ref{eq:particle_velocity}}
\mathbf{v}_p^{n+1} = \sum_{i=1}^{N} w_{ip}^n \mathbf{v}_i^{n+1}
\end{equation}

If the external body force is applied to particles rather than to grid nodes, in each iteration, we need to transfer its impact to the grid nodes (i.e. particle to grid, \textit{P2G}):

\begin{equation}
\mathbf{f}_i^{ext} = \sum_{p=1}^{M} w_{ip}^n \mathbf{f}_p^{ext}
\end{equation}

\noindent and then apply the combined force to update grid velocities using Eq.\ref{eq:Eulerian_time_integration}.

\paragraph{Evolving the IPS} We have now prepared the basic ingredients for simulating an IPS. To evolve the particles, we solve the momentum equation at grid nodes at discrete times, which means in each iteration, as the particles move, particle quantities such as particle velocity $\mathbf{v}_p$, particle position $\mathbf{x}_p$, the deformation gradient matrix $\mathbf{F}_p$, as well as the APIC transfer tensors $\mathbf{B}_p$ and $\mathbf{D}_p$ (if APIC rather than PIC transfer is used), and nodal quantities such as nodal mass $m_i$, nodal momentum $m_i \mathbf{v}_i$, velocity $\mathbf{v}_i$, internal nodal force $\mathbf{f}_i^{int}$, as well as the interpolation weights $w_{ip}$, are updated. Note that while nodal masses $m_i$ are dynamically changing, particle masses $m_p$ are never changed (mass preserving). Also no need to update nodal positions $\mathbf{x}_i$.

We start with $M$ initialised particles and $N$ grid nodes, each of them are assigned with the aforementioned properties. Assuming we are at iteration $t=n$, we first prepare the interpolation weights $w_{ip}^n$ and weight gradients $\nabla w_{ip}^n$ using current grid position $\mathbf{x}_i$ and particle position $\mathbf{x}_p^n$ (Eq.\ref{eq:interpolation_weights} and Eq.\ref{eq:interpolation_weight_gradients}). Both values are stored on particles for later property states transfer. Then we transfer particle states to grid nodes (\textit{particle-to-grid}, P2G), i.e. calculating grid node mass $m_i^n$ and momentum $m_i^n \mathbf{v}_i^n$ via the \textit{particle-in-cell} (PIC) \cite{Jiang2015thesis} or \textit{affine particle-in-cell} (APIC) \cite{Jiang2015} routine - both transfer methods map the particle mass $m_p$ and momentum $m_p \mathbf{v}_p$ to (nearby) nodes:

\begin{equation} \label{eq:cc_PIC} \tag{cc.Eq.\ref{eq:PIC}}
    \text{\textit{PIC}:} \quad m_i^n = \sum_{p=1}^{M} w_{ip}^n m_p, \quad m_i^n \mathbf{v}_i^n = \sum_{p=1}^{M} w_{ip}^n m_p \mathbf{v}_p^n
\end{equation}

\begin{equation} \label{eq:cc_APIC} \tag{cc.Eq.\ref{eq:APIC}}
    \text{\textit{APIC}:} \quad m_i^n = \sum_{p=1}^{M} w_{ip}^n m_p, \quad m_i^n \mathbf{v}_i^n = \sum_{p=1}^{M} w_{ip}^n m_p [\mathbf{v}_p^n + \mathbf{B}_p^n (\mathbf{D}_p^n)^{-1} (\mathbf{x}_i - \mathbf{x}_p^n)]
\end{equation}

\noindent where $\mathbf{D}_p^n$ and $\mathbf{B}_p^n$ are tracked (only in APIC) as per: 

\begin{equation*}
    \mathbf{D}_p^n= \sum_{i=1}^{N} w_{ip}^n (\mathbf{x}_i - \mathbf{x}_p^n) (\mathbf{x}_i - \mathbf{x}_p^n)^\top \quad \text{(cc.Eq.\ref{eq:APIC_Dp})}, \quad 
    \mathbf{B}_p^{n} = \sum_{i=1}^{N} w_{ip}^{n-1} \mathbf{v}_i^{n} (\mathbf{x}_i - \mathbf{x}_p^{n-1})^\top \quad \text{(cc.Eq.\ref{eq:APIC_Bp})} \\
\end{equation*}

\noindent Grid node velocities $\mathbf{v}_i$ can be be calculated using the aggregated nodal momentum:

\begin{equation} \label{eq:grid_velocity}
    \mathbf{v}_i^n = \frac{m_i^n \mathbf{v}_i^n}{m_i^n} 
\end{equation}

\noindent and internal nodal forces calculated by aggregating all stress contributions induced by the deformation gradients at particles:

\begin{equation} \label{eq:cc_MPM_grid_forces_energy_density} \tag{cc.Eq.\ref{eq:MPM_grid_forces_energy_density}}
    \mathbf{f}_i^{int,n} = \mathbf{f}_i^{int}(\mathbf{x}_i) = - \sum_{p=1}^{M} V_p^0 \left[\frac{\partial \Psi(\mathbf{F}_p^n)}{\partial \mathbf{F}} \right] (\mathbf{F}_p^n)^\top \nabla w_{ip}^n
\end{equation}

\noindent Summing up all internal and external forces, we can follow \textit{Newton's second law} (a.k.a the momentum equation) to update nodal velocities:

\begin{equation*} \tag{cc.Eq.\ref{eq:Eulerian_time_integration}}
    \mathbf{v}_i^{n+1} = \mathbf{v}_i^n + \Delta t \cdot (\mathbf{f}_i^{int,n} + \mathbf{f}_i^{ext}) / m_i 
\end{equation*}

\noindent the updated nodal velocities are then propagated back to particles (\textit{grid-to-particle}, G2P) for updating particle property states (Eq.\ref{eq:particle_velocity}, Eq.\ref{eq:APIC_Bp}, Eq.\ref{eq:MPM_deformaiton_gradients_update}):

\begin{equation*}
\begin{aligned}
    & \mathbf{v}_p^{n+1} = \sum_{i=1}^{N} w_{ip}^{n} \mathbf{v}_i^{n+1}, \quad \mathbf{x}_p^{n+1} = \mathbf{x}_p^{n} + \Delta t \cdot \mathbf{v}_p^{n+1} \\
    & \mathbf{B}_p^{n+1} = \sum_{i=1}^{N} w_{ip}^{n} \mathbf{v}_i^{n+1} (\mathbf{x}_i - \mathbf{x}_p^{n})^\top, \quad \mathbf{F}_p^{n+1} = \left( \mathbf{I} + \Delta t \cdot \sum_{i=1}^{N} \mathbf{v}_i^{n+1} (\nabla w_{ip}^n)^\top \right) \mathbf{F}_p^n
\end{aligned}
\end{equation*}

\begin{figure}
    \centering
    \includegraphics[width=0.9\linewidth]{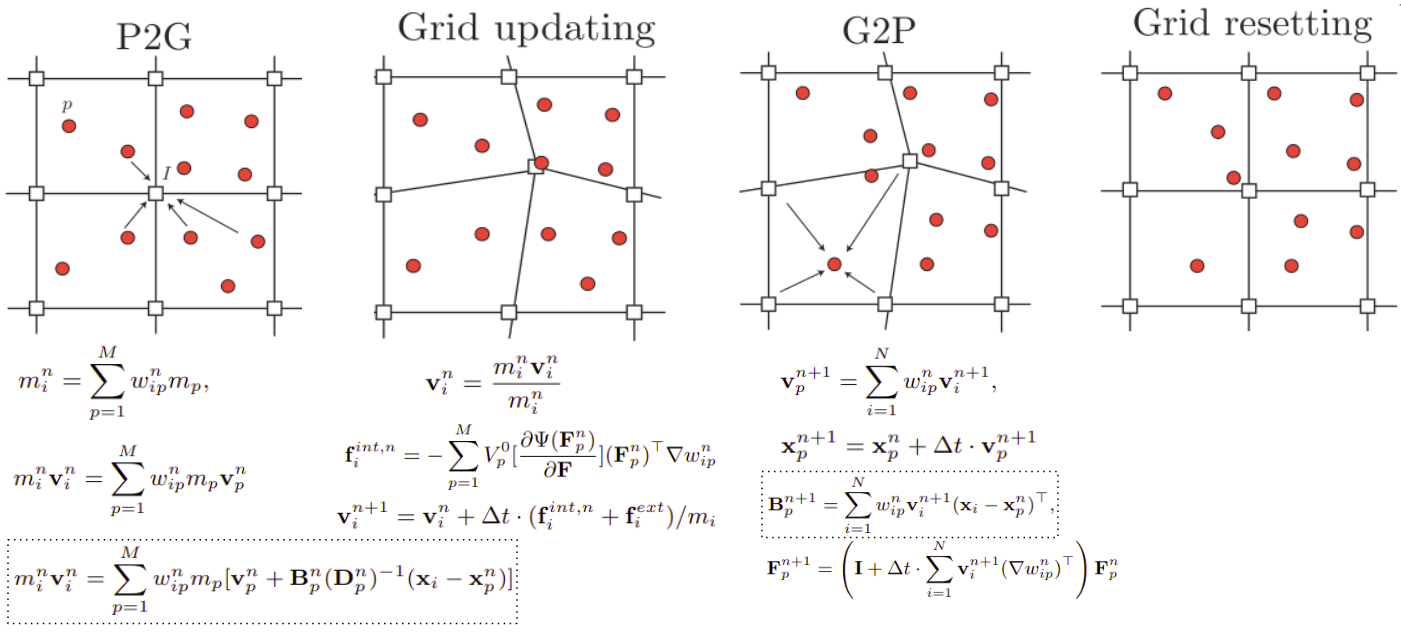}
    \caption{One MPM-ParVI cycle illustrating Algo.\ref{algo:MPM_sampling1} and Algo.\ref{algo:MPM_sampling2}. Red circles represent material points overlaid on a regular mesh grid with nodes represented by small squares. \textit{P2G}: arrows represent particle property states (mass and momentum) being projected (aggregated) onto grid nodes. \textit{Grid updating}: equations of motion are solved on grid nodes, resulting in updated nodal velocities. \textit{G2P}: arrows represent the updated nodal property states (nodal velocities) being mapped back to particles. Finally, grid is reset. Figures are modified from \cite{DeVaucorbeil2020}, equations within dashed boxes are only used in Algo.\ref{algo:MPM_sampling2}.}
    \label{fig:MPM_procedure}
\end{figure}

The procedure of a basic MPM-ParVI cycle, as illustrated above, is shown in Fig.\ref{fig:MPM_procedure}. This MPM-ParVI procedure is the same as a classic MPM procedure \cite{Jiang2015thesis} except that external effects, represented by $\mathbf{f}_i^{ext}(\mathbf{x}_i)=\nabla_{\mathbf{x}} \log p(\mathbf{x}_i)$, are inserted when updating nodal velocities. The MPM scheme used combines the basic implementations from \cite{Jiang2015thesis} and \cite{Jiang2016}, with some modifications though. Each cycle consists of 4 steps: (1) \textbf{Particle to grid} (P2G). Information of particles, e.g. particle masses $m_p$ and velocities $\mathbf{v}_p$, are transferred to background grid nodes. (2) \textbf{Grid updating}. Calculate nodal velocities $\mathbf{v}_i$, internal nodal forces \footnote{Note that, calculating the internal nodal forces $\mathbf{f}_i^{int}$ from particle deformation gradients $\mathbf{F}_p$ can also be seen as a P2G step.} $\mathbf{f}_i^{int}$, solve discrete momentum equations and update nodal velocities $\mathbf{v}_i$. (3) \textbf{Grid to particles} (G2P). Information of the updated nodes, e.g. nodal velocities $\mathbf{v}_i$, are transferred back to particles for updating particle positions $\mathbf{x}_p$, velocities $\mathbf{v}_p$, deformation gradients $\mathbf{F}_p$, affine matrices $\mathbf{B}_p$, etc. (4) \textbf{Grid resetting}. Grid is reset to its original state or a new grid is used (every several iterations), therefore mesh distortion never happens (perfect for large deformation simulation). 

We keep evolving the particle system until an optimal configuration is achieved. This optimal configuration can be a transient or steady state which best approximates the target shape. We can define certain metrics, e.g. average particle density or overall probability (or other proper clustering indicator), to be monitored throughout the evolution process to find a terminal state. The MPM-ParVI algorithm is presented in Algo.\ref{algo:MPM_sampling1}. A slightly different variant, using the APIC method \cite{Jiang2015} for P2G transfer, is presented in Algo.\ref{algo:MPM_sampling2}.

\begin{algorithm}
\fontsize{8}{8}
\small
\caption{MPM-based sampling (PIC transfer)}
\label{algo:MPM_sampling1}
\begin{itemize}
\item{\textbf{Inputs}: a differentiable target density $p(\mathbf{x})$ with magnitude amplification constant $\alpha$, total number of particles $M$, particle (density) dimension $d$, initial proposal distribution $p^0(\mathbf{x})$, total number of grid nodes $N$, total number of iterations $T$, step size $\Delta t$. An energy density function $\Psi$ specifying the constitutive model, and an interpolation function $K$.} 
\item{\textbf{Outputs}: Particles located at positions $\{\mathbf{x}_p \in \mathbb{R}^d \}_{p=1}^{M}$ whose empirical distribution approximates the target density $p(\mathbf{x})$.}
\end{itemize}
\vskip 0.06in
1. \textbf{\textit{Initialise}} particles and grid. [$\mathcal{O}(max(M,N)d)$]
    \begin{addmargin}[1em]{0em}% 
    Initialise $M$ particles from $p^0(\mathbf{x})$, each with mass (constant) $m_p$, volume $V_p^0$, initial position $\mathbf{x}_p$, initial velocity $\mathbf{v}_p$, and material-related parameters. Some default values can be set to be 0 \cite{Jiang2015thesis}. [$\mathcal{O}(Md)$] \\ 
    Initialise a grid with $N$ nodes. Grid node locations $\mathbf{x}_i, i=1,2,...,N$ are on a undeformed regular grid and are time-invariant. Initialise grid data, i.e. nodal mass $m_i$, velocity $\mathbf{v}_i$, to default values of 0. [$\mathcal{O}(Nd)$]
    \end{addmargin}

2. \textbf{\textit{Update}} particle positions. \\
For each iteration $t=1,2,...,T$, repeat until optimal configuration:
    \begin{addmargin}[1em]{0em}% 
    (1) Compute the interpolation weights $w_{ip}^n$ and weight gradients $\nabla w_{ip}^n$ for each particle $p$:
    \[
    w_{ip}^n=K(\mathbf{x}_i - \mathbf{x}_p^n), \quad \nabla w_{ip}^n=\nabla K(\mathbf{x}_i - \mathbf{x}_p^n)
    \]
    These location-dependent weights and weight gradients are computed once and stored on the particles in each iteration. \\
    (2) Transfer \textbf{particle} quantities to the \textbf{grid} (\textbf{\textit{P2G}}). Compute the \textbf{grid} mass and momentum using PIC (Eq.\ref{eq:PIC}): [$\mathcal{O}(MNd)$]
    \[
    m_i^n = \sum_{p=1}^{M} w_{ip}^n m_p, \quad m_i^n \mathbf{v}_i^n = \sum_{p=1}^{M} w_{ip}^n m_p \mathbf{v}_p^n
    \]
    (3) Compute \textbf{grid} velocities (Eq.\ref{eq:grid_velocity}): [$\mathcal{O}(Nd)$]
    \[
   \mathbf{v}_i^n = \frac{m_i^n \mathbf{v}_i^n}{m_i^n} 
    \]
    (4) Compute explicit \textbf{grid} forces $\mathbf{f}_i^{int,n}$ (Eq.\ref{eq:MPM_grid_forces_energy_density}): [$\mathcal{O}(Md^3+MNd^2)$] \\
    \[
    \mathbf{f}_i^{int,n} = - \sum_{p=1}^{M} V_p^0 [\frac{\partial \Psi(\mathbf{F}_p^n)}{\partial \mathbf{F}} ] (\mathbf{F}_p^n)^\top \nabla w_{ip}^n
    \]
    \\ 
    (5) Update \textbf{grid} velocities $\mathbf{v}_i^{n}$ using a explicit Euler time integration scheme \footnotemark[1] (Eq.\ref{eq:Eulerian_time_integration}): [$\mathcal{O}(Nd)$] \\
    \[
    \mathbf{v}_i^{n+1} = \mathbf{v}_i^n + \Delta t \cdot (\mathbf{f}_i^{int,n} + \mathbf{f}_i^{ext}) / m_i
    \]
    where the time-invariant, external body force $\mathbf{f}_i^{ext} = \alpha \nabla_{\mathbf{x}} \log p(\mathbf{x}_i)$. \\
    (6) Transfer \textbf{grid} property states to \textbf{particles} (\textbf{\textit{G2P}}). Compute \footnotemark[2] the new \textbf{particle} velocities $\mathbf{v}_p^{n+1}$ (Eq.\ref{eq:particle_velocity}) and positions $\mathbf{x}_p^{n+1}$; update the deformation gradient matrix $\mathbf{F}_p^{n}$ (Eq.\ref{eq:MPM_deformaiton_gradients_update}): [$\mathcal{O}(Md^3+MNd^2)$] \\
    \[
    \mathbf{v}_p^{n+1} = \sum_{i=1}^{N} w_{ip}^{n} \mathbf{v}_i^{n+1}, \quad \mathbf{x}_p^{n+1} = \mathbf{x}_p^{n} + \Delta t \cdot \mathbf{v}_p^{n+1}, \quad \mathbf{F}_p^{n+1} = \left( \mathbf{I} + \Delta t \cdot \sum_{i=1}^{N} \mathbf{v}_i^{n+1} (\nabla w_{ip}^n)^\top \right) \mathbf{F}_p^n
    \]
    (7) Re-set grid data, i.e. nodal mass $m_i$, velocity $\mathbf{v}_i$, to the default values of 0.
    \end{addmargin}

3. \textbf{\textit{Return}} the intermediate or final particle positions $\{\mathbf{x}_p^{T}\}_{p=1}^{M}$ and their histogram and/or \textit{KDE} estimate for each dimension. \\
\end{algorithm}
\footnotetext[1]{Explicit Euler time integration is easy to implement but requires smaller time step sizes; implicit backward Euler time integration requires solving a linear system to obtain $\mathbf{v}_i^{n+1}$ but allows for larger time steps \cite{Jiang2015thesis}. Note that we don't need to update grid node positions $\mathbf{x}_i$.}
\footnotetext[2]{An alternative, hybrid PIC and FLIP scheme for updating the particle velocity and position is provided by \cite{Jiang2015thesis} (pp.12): $\mathbf{v}_p^{n+1} = (1 - \alpha) \sum_i \mathbf{v}_i^{n+1} w_{ip}^n + \alpha \left( \mathbf{v}_p^n + \sum_i (\mathbf{v}_i^{n+1} - \mathbf{v}_i^n) w_{ip}^n \right)$ and $\mathbf{x}_p^{n+1} = \mathbf{x}_p^n + \Delta t \sum_i \mathbf{v}_i^{n+1} w_{ip}^n$.}

\section{Discussions} \label{sec:discussions}

As the MPM-ParVI method is built on classic MPM, here we discuss some topics related to both.

\paragraph{Lagrangian and Eulerian descriptions of motion} Both Eulerian (spatial) and Lagrangian (material) descriptions are used in solid and fluid mechanics to describe material deformation or flow. Lagrangian description assumes observer flows with the observation element (e.g. the particle).  The grid, if any, is attached to the element and therefore deforms (can be distorted) during the simulation process. This facilitates tracking of element deformation history. As computation is mainly performed on particles, Lagrangian methods (e.g. SPH) are very efficient. Eulerian description, on the other hand, issues a fixed grid in space and material flows through the grid. It avoids grid distortion but induces expensive computation. Lagrangian description focuses on individual particles, it uses the initial configuration (i.e. $t$=0) to describe the physical quantities and the deformation state of a continuum body. The Eulerian description focuses on a specific position in space at current time $t$, it describes the motion of a continuum body with respect to the current coordinates and time \cite{Kumar2022}. MPM takes a hybrid Lagrangian and Eulerian view: dynamics are solved on the computational Eulerian grid while particles act as quadrature points \cite{Jiang2016}. For a comprehensive read about the Lagrangian and Eulerian descriptions of fluids, see e.g. \cite{Talpaert2002}.

\paragraph{Time step size} MPM allows for larger time step size $\Delta t$ than other particle-based simulation methods such as SPH, as the $\Delta t$ in MPM relies on the cell size (i.e. the grid spacing) rather than the particle space (as in SPH) \cite{DeVaucorbeil2020}. Both Algo.\ref{algo:MPM_sampling1} and Algo.\ref{algo:MPM_sampling2} use explicit time integration which is easy to implement but requires smaller $\Delta t$. Implicit time integration \cite{Stomakhin2013} requires solving a linear system but allows for larger $\Delta t$ \cite{Jiang2015thesis}.

\paragraph{Multi-modal inference} Inference of multi-modal densities require particles to separate and form local groups, which is similar to fracture or cracks modelling using MPM: there are multiple internal surfaces across which the displacement field is discontinuous, and multiple velocity fields at nodes whose supports are cut by the cracks. MPM is efficient for modelling problems with moving discontinuities such as fracture evolution \cite{Perez2021}, and well suited for large motion and shape change problems.

\paragraph{Memory and complexity} In $d$-dimensional space, the deformation gradient matrix $\mathbf{F}_p$, the 1st PK stress matrix $\mathbf{P}$, the APIC tensors $\mathbf{B}_p$ and $\mathbf{D}_p$, are all of dimension $d \times d$. One need to store $d^2$ numbers for each matrix at each particle. In addition, each vector-valued particle and nodal quantity, e.g. position, velocity, momentum and force, requires $d$ numbers to be stored. If we use a regular Cartesian grid and assign $k$ nodes along each dimension, in total we have $N=k^d$ nodes. In each iteration $n$, we need $\mathcal{O}((N+M)d+Md^2)$ space for storing these vectors and matrices.

The time cost for each MPM step has been attached in each step in both algorithms, taking into account matrix inversion, transpose, matrix-matrix and matrix-vector multiplications \footnote{For matrices of dimension $d \times d$ and vector of length $d$, matrix inversion and multiplication take $d^3$, matrix-vector multiplication takes $d^2$, vector inner product takes $d$, vector out product takes $d^2$, scalar-vector product takes $d$, all in FLOPS. Singular value decomposition of the deformation gradient matrix $\mathbf{F}$ is considered in e.g. \cite{Jiang2015,DeVaucorbeil2020}.}. Overall, the MPM algorithm takes \footnote{Note in Algo.\ref{algo:MPM_sampling2}, in each iteration, for step 2.1, we pre-calculate $\mathbf{B}_p^n (\mathbf{D}_p^n)^{-1}$ and cache it, which takes $\mathcal{O}(Md^3+MNd^2)$. If not caching, the cost will be $\mathcal{O}(MNd^3)$. The same applies to $[\frac{\partial \Psi(\mathbf{F}_p^n)}{\partial \mathbf{F}} ] (\mathbf{F}_p^n)^\top$ in step 2.3.} $\mathcal{O}(Md^2(d+N))$, which is about $\mathcal{O}(MNd^2)$ as normally we have $d \ll N$. It is smaller than $\mathcal{O}(MNd^3)$ but larger than $\mathcal{O}(MNd)$. For low dimensions (i.e. $d$=1,2,3), normally we have $M>N$; in high-dimensions, however, $M \ll N=k^d$.

The bottleneck therefore lies in the two multiplicative factors $N=k^d$ and $d^2$. It is suspicious to say that, by modelling indirect particle interactions, MPM reduces the computational burden from $\mathcal{O}(M^2)$ to $\mathcal{O}(MN)$.

\paragraph{Advantages of MPM-ParVI} (1) \textbf{Advantages of MPM}. MPM provides a particle simulation framework which enjoys the advantages of both Lagrangian and Eulerian descriptions \cite{Perez2021}: no element distortion happens and no remeshing is required during simulation. The Lagrangian formulation allows particle motion and material deformation histories to be tracked throughout the simulation process, while the Eulerian grid enables automatic treatment of self-collision and fracture \cite{Jiang2016}. Further, it is free of mesh-entanglement problems, and allows for both explicit \cite{Jiang2015} and implicit \footnote{MPM allows a grid-based implicit integration scheme which is independent of the number of Lagrangian particle \cite{Jiang2016}.} \cite{Jiang2016,Zheng2022,Zheng2022b} time integration schemes. More advantages of MPM are listed in e.g. \cite{DeVaucorbeil2020}. (2) \textbf{Physics-based and deterministic}. MPM-ParVI follows deterministic physics to generate samples, the results are trackable and reproducible. Interactions between particles are modelled in a principled manner which preserves mass, momentum and energy. (3) \textbf{Handling complex geometries}. As MPM is efficient in modelling discontinuities, large motion and shape change problems, MPM-ParVI may be of advantage for inferring complex (e.g. multi-modal) densities. (4) \textbf{Ease of implementation}. MPM-ParVI has few hyper-parameters to tune (typically, the mesh spacing and time step) \footnote{However, one do need to choose a proper constitutive model to ensure alignment between the inference and simulation tasks.}, which is convenient as compared to SPH, and it can be easily implemented for parallel and distributed computing. (5) \textbf{Speed and accuracy}. It is reported \cite{Ma2009b,DeVaucorbeil2020} that, MPM is faster and more accurate than SPH due to larger time step size and no neighbourhood searching. However, as a new class of ParVI methods, comparison of these physics-based ParVI methods with other ParVI/VI methods has not been studied.

\paragraph{Limitations of MPM-ParVI} (1) \textbf{Disadvantages of MPM}. MPM is generally less accurate than FEM. Particles in MPM can suffer from numerical fracture, grid-crossing instability \footnote{\textit{Grid-crossing error} is one of the well-known shortcomings of MPM, which is induced by (insufficient) smoothness of the interpolation shape function \cite{Perez2021}. If a shape function with $C^0$ degree of smoothness (first-order derivative is discontinuous between cells) is employed, spurious variations (e.g. jumps in the stress) can result when particles cross cell boundaries \cite{Perez2021}. Remedies have been proposed, e.g. GIMP \cite{Bardenhagen2004,Ma2006}, DDMPM \cite{Zhang2011}, CPDI \cite{Sadeghirad2011}, IGA-MPM \cite{Moutsanidis2020}, LME-MPM \cite{Perez2021}.}, the null space issue and no convergence for very fine meshes \cite{DeVaucorbeil2020}. Large memory is required for computation at grid and particles. Analysis of convergence, error and stability is challenging \cite{DeVaucorbeil2020}. The convergence rate is rarely of second order \cite{DeVaucorbeil2020}. Boundary conditions are difficult to enforced. More disadvantages of MPM are listed in e.g. \cite{DeVaucorbeil2020}. (2) \textbf{Reduced flexibility} as compared to SPH due to the hybrid Lagrangian-Eulerian nature of MPM. Although both SPH and MPM requires choosing a proper kernel (for different purposes), in MPM the modeller also has to choose a constitutive model (e.g. the energy density function $\Psi$) for simulation, and the choice of which can have impact on the inference results. There is some flexibility though, for devising MPM-ParVI. For example, one can apply a time variant, external body force, in an annealing style, to control convergence. Also, one may formulate the information from the target density as boundary condition(s). (3) \textbf{Limited scalability}. MPM was developed for real-world simulations which are typically low-dimensional. Statistical inference problems, however, can range from low to high dimensions, and MPM-ParVI suffers from curse of dimensionality as a background mesh grid is needed for the simulation. The computational efforts increases as $\mathcal{O}(Md^3+MNd^2)$ where $N=k^d$, $k$ being the number of grid nodes along each dimension. This intrinsic exponential growth may prevent it from being applied to high dimensions.

\paragraph{Physical statistics} This work extends the author's previous work on physics-based ParVI methods \cite{huang2024EParVI,huang2024SPHParVI}, which forms a new branch of \textit{physical statistics} (against the long standing statistical physics which applies statistics to physics) and coins the term \textit{science for AI} (against the trending \textit{AI for science}). There have been many physically and biologically inspired AI algorithms, e.g. the genetic algorithm \cite{Holland1975}, particle swarm optimization (PSO \cite{Kennedy1995,YANG2020101104}) and HMC. This series of work, however, is first of this kind which formulates a statistical inference problem fully as a physical simulation process. Unlike traditional ParVI methods which use statistical and optimisation principles to perform inference, these new ParVI methods are purely based on physics \footnote{By following the physical dynamics, these physical simulation processes are, of course, (implicitly) optimizing certain objective functions such as potential or energy (e.g. principle of least action). This provides a future direction for proving existence and potentially uniqueness of the optimal solution.}. SPH-ParVI, for example, employs the Navier–Stokes equations, which represent mass, momentum and energy conservation, as deterministic dynamics to guide the inference procedure. EParVI, on the other hand, utilizes principles of electrostatics to model the IPS. 

\paragraph{Applications} Similar to other gradient-based sampling methods such as Hamiltonian Monte Carlo (HMC), Langevin Monte Carlo (LMC), SVGD, and SPH-ParVI \footnote{SPH-ParVI can perform sampling either using directly the known-up-to-a-constant density, or using gradients \cite{huang2024EParVI}.}, MPM-ParVI can be used to infer partially known densities, e.g. an intractable Bayesian posterior known up to the normalising constant. By plugging the gradient of log \textit{pdf}, i.e. the \textit{score} which eliminates the unknown constant, into the MPM formulation, one can recover the target density approximately. Also, It can be used to sample score-based generative models \cite{SM_Yang,SBGM_Huang} in which the score can be empirically estimated via \textit{score-matching} \cite{hyvarinen2005sm}.

\section{Conclusion and future work}

\paragraph{Conclusion} 

We propose the MPM-ParVI algorithm for sampling and variational inference. It formulates sampling fully as a physical simulation process. By incorporating the gradient field of a target density into the MPM method, it drives a chosen material, represented as a system of interacting particles, to deform towards a geometry which approximates the target distribution. Particle motions are governed by principled, deterministic physics, concentration and diversity of particles are naturally maintained by the MPM formulation. This easy-to-implement ParVI method offers deterministic sampling and inference for a class of probabilistic models such as those encountered in Bayesian inference (e.g. intractable densities) and generative modelling (e.g. score-based). As a hybrid Lagrangian-Eulerian method, it suffers from curse of dimensionality.

\paragraph{Future work} There are active innovations to enhance the capability (e.g. efficiency, stability, etc) and applicability \cite{He2023,gmd2020} of MPM. The bottleneck for scaling MPM-ParVI to high dimensions lies in the exponential growth of computational efforts \textit{w.r.t} the dimension $d$, therefore improving the scalability of MPM-ParVI may deserve some future attention. Also, formal analysis of the physics-guided particle evolution from an optimisation perspective may be a future direction, as well as theoretically proving the existence and potentially uniqueness of the optimal solution. Undergoing work includes experimental validation of the MPM-ParVI method, particularly its efficacy in inferring complex densities.

\bibliography{reference}

\appendix

\section{Abbreviations} \label{app:abbreviations}

Abbreviations used in this context are listed in Table.\ref{tab:terminologies}.

\begin{table}[h!]
\centering
\begin{tabular}{ll}
\toprule
\textbf{Abbreviation} & \textbf{Full name} \\
\midrule
1st PK & first Piola-Kirchoff stress \\
2nd PK & second Piola-Kirchoff stress \\
APIC & Affine Particle-In-Cell \\
BSMPM & B-splines Material Point Method \\
CPDI & Convected Particle Domain Integrator \\
DDMPM & Dual Domain Material Point Method \\
DEM & Discrete Element Method or Diffuse Element Method \\
DGMPM & Discontinuous Galerkin Material Point Method \\
EFG & Element Free Galerkin method \\
ELBO & Evidence Lower Bound \\
EoS & Equation of State \\
EParVI & Electrostatics-based ParVI \\
EVI & Energetic Variational Inference \\
FDM & Finite Difference Method \\
FEM & Finite Element Method \\
FLIP & Fluid Implicit Particle method \\
FVM & Finite Volume Method \\
G2P & Grid-to-Particle \\
GIMP & Generalized Interpolation Material Point \\
HMC & Hamiltonian Monte Carlo \\
IGA-MPM & Isogeometric Material Point Method \\
iMPM & improved Material Point Method \\
IPS & Interacting Particle System \\
LMC & Langevin Monte Carlo \\
LME-MPM & Local Maximum Entropy Material Point Method \\
MLPG & Meshless Local Petrov Galerkin method \\
MCMC & Markov chain Monte Carlo \\
MPFEM & Material Point Finite Element Method \\
MPM & Material Point Method \\
MPM-ParVI & MPM-based ParVI \\
OTM & Optimal Transport Meshfree method \\
P2G & Particle-to-Grid \\
ParVI & Particle-based Variational Inference \\
PFEM & Particle Finite Element Method \\
PIC & Particle-In-Cell \\
PSO & Particle Swarm Optimization \\
RKPM & Reproducing Kernel Particle Method \\
SMC & Sequential Monte Carlo \\
SPH & Smoothed Particle Hydrodynamics \\
SPH-ParVI & SPH-based ParVI \\
SVGD & Stein Variational Gradient Descent \\
TLMPM & Total Lagrangian Material Point Method \\
\bottomrule
\end{tabular}
\caption{Terminologies and abbreviations.}
\label{tab:terminologies}
\end{table}

\section{Quantities and symbols} \label{app:quantities_MPM}

Quantities and symbols used in this context are listed in Table.\ref{table:mpm_quantities}.

\begin{table}[h!]
\small
\centering
\begin{tabular}{| m{6.2cm} | m{2.5cm} | m{2cm} | m{1.5cm} |}
\hline
\textbf{Name} & \textbf{Symbol} & \textbf{Defined or stored on} & \textbf{Type} \\
\hline
1st PK stress tensor & $\mathbf{P}$ & Grid & Tensor \\
\hline
2nd PK stress tensor & $\mathbf{S}$ & Grid & Tensor \\
\hline
Coordinate mapping function & $\phi$ & - & - \\
\hline
Density & $\rho$ & - & Scalar \\
\hline
Dimension & $d$ & - & Scalar \\
\hline
Distance & $r=(x_p-x_i)/h$ & Particle & Scalar \\
\hline
Divergence & $\nabla \cdot$ & - & - \\
\hline
Energy density function & $\Psi$ & - & Scalar \\
\hline
First Lame parameter & $\mu$ & - & Scalar \\
\hline
Gradient & $\nabla$ & - & - \\
\hline
Gravitational acceleration  & $\mathbf{g}$ & - & Vector \\
\hline
Grid spacing & $h$ & - & Scalar \\
\hline
Identity matrix & $\bm{I}$ & - & Matrix \\
\hline
Integration time step size & $\Delta t$ & - & Scalar \\
\hline
Interpolation kernel & $K$ & - & Scalar \\
\hline
Interpolation weight & $w_{ip}$ & Particle & Scalar \\
\hline
Interpolation weight gradient & $\nabla w_{ip}$ & Particle & Vector \\
\hline
Iteration index & $t$ & - & Scalar \\
\hline
Iteration no. & $n$ & - & Scalar \\
\hline
Kirchhoff stress & $\boldsymbol{\tau}$ & Grid & Tensor \\
\hline
Material derivative & $D/Dt$ & - & - \\
\hline
Material domain & $\Omega$ & - & - \\
\hline
Node external body force & $\mathbf{f}^{ext}_i$ & Grid & Vector \\
\hline
Node internal force & $\mathbf{f}_i^{int}$ & Grid & Vector \\
\hline
Node mass & $m_i$ & Grid & Scalar \\
\hline
Node momentum & $m_i \mathbf{v}_i$ & Grid & Vector \\
\hline
Node position (Eulerian) & $\mathbf{x}_i$ & Grid & Vector \\
\hline
Node velocity & $\mathbf{v}_i$ & Grid & Vector \\
\hline
Particle acceleration & $\mathbf{a}_p$ & Particle & Vector \\
\hline
Particle APIC tensor & $\mathbf{D}_p$ & Particle & Tensor \\
\hline
Particle APIC tensor (particle affine matrix) & $\mathbf{B}_p$ & Particle & Tensor \\
\hline
Particle Cauchy stress tensor & $\boldsymbol{\sigma}_p$ & Particle & Tensor \\
\hline
Particle deformation gradient & $\mathbf{F}_p$, $ J=1/det(\mathbf{F})$ & Particle & Tensor \\
\hline
Particle displacement & $\mathbf{u}_p$ & Particle & Vector \\
\hline
Particle mass & $m_p$ & Particle & Scalar \\
\hline
Particle momentum & $m_p \mathbf{v}_p$ & Particle & Vector \\
\hline
Particle or pdf & $p$ & - & - \\
\hline
Particle position (Eulerian) & $\mathbf{x}_p$ & Particle & Vector \\
\hline
Particle position (Lagrangian) & $\mathbf{X}_p$ & Particle & Vector \\
\hline
Particle volume & $V_p$ & Particle & Scalar \\
\hline
Particle velocity & $\mathbf{v}_p$ & Particle & Vector \\
\hline
pdf amplification constant  & $\alpha$ & - & Scalar \\
\hline
Poisson's ratio & $\nu$ & - & Scalar \\
\hline
Second Lame parameter & $\lambda$ & - & Scalar \\
\hline
Strain & $\boldsymbol\epsilon$ & Grid & Tensor \\
\hline
Summation & $\sum$ & - & - \\
\hline
Total no. of iterations & $T$ & - & Scalar \\
\hline
Total number of nodes & $N$ & - & Scalar \\
\hline
Total number of particles & $M$ & - & Scalar \\
\hline
Young’s modulus & $E$ & - & Scalar \\
\hline
\end{tabular}
\caption{Quantities and symbols used.}
\label{table:mpm_quantities}
\end{table}

\section{MPM: more details} \label{app:MPM_more_details}

We mainly follow \cite{DeVaucorbeil2020}, \cite{Jiang2015thesis} and \cite{Jiang2016}, with small modifications, to present more details on deriving MPM. All vectors are column vectors by default. 

\paragraph{Background of continuum mechanics} Here we mainly follow pp.23-26 in \cite{DeVaucorbeil2020}. In continuum mechanics, the \textit{displacement} $\mathbf{u}$, \textit{velocity} $\mathbf{v}$ and \textit{acceleration} $\mathbf{a}$ are used to describe the motion of a body. These quantities are time and position dependent, we first note \footnote{$\mathbf{X}$ is defined on the \textit{Lagrangian or material coordinate}, $\mathbf{x}$ is defined on the \textit{Eulerian or spatial coordinate \cite{DeVaucorbeil2020}.}. Normally a regular Cartesian grid is used.} the relation between the Lagrangian coordinate $\mathbf{X}$ and Eulerian coordinate $\mathbf{x}$:

\begin{equation} \label{eq:Lagrangian_Eulerian_coordinates}
    \mathbf{x} = \phi (\mathbf{X},t)
\end{equation}

\noindent where $\phi$ is a mapping function.

The displacement $\mathbf{u}(\mathbf{X},t)$ of a material point $\mathbf{X}$ is the difference between its current position $\phi(\mathbf{X},t)$ and initial position $\phi(\mathbf{X},0)$:

\begin{equation} \label{eq:displacement}
    \mathbf{u}(\mathbf{X},t) := \phi(\mathbf{X},t) - \phi(\mathbf{X},0) = \mathbf{x} - \mathbf{X}
\end{equation}

The Lagrangian velocity $\mathbf{v}(\mathbf{X},t)$ of a material point $\mathbf{X}$ is:

\begin{equation} \label{eq:velocity}
    \mathbf{v}(\mathbf{X},t) := \frac{\partial \phi (\mathbf{X},t)}{\partial t}
\end{equation}

The acceleration $\mathbf{a}(\mathbf{X},t)$ of a material point $\mathbf{X}$ is:

\begin{equation} \label{eq:acceleration}
    \mathbf{a}(\mathbf{X},t) := \frac{\partial \mathbf{v} (\mathbf{X},t)}{\partial t} = \frac{\partial^2 \phi (\mathbf{X},t)}{\partial t^2}
\end{equation}

The \textit{deformation gradient tensor} $\mathbf{F}$, which maps the undeformed configuration to the deformed configuration, is defined as:

\begin{equation} \label{eq:deformation_gradient}
    \mathbf{F}(\mathbf{X},t) := \frac{\partial \phi(\mathbf{X},t)}{\partial \mathbf{X}} = \frac{\partial \mathbf{x}(\mathbf{X},t)}{\partial \mathbf{X}}.
\end{equation}

\noindent In 3D space, for example, the deformation gradient tensor $\mathbf{F}$ is a $3 \times 3$ matrix with components describing how the material moves and deforms, accounting for stretching, shearing, and rotation of material elements. $\mathbf{F}$ is key in deriving all other deformation quantities. If we denote $J=1/\det(\mathbf{F})$, some useful derivations are \cite{Jiang2015}:

\begin{equation}
\begin{aligned}
    & \frac{\partial J}{\partial \mathbf{F}} = J \mathbf{F}^{-\top} \\
    & \mathbf{F}=\mathbf{U} \boldsymbol{\Sigma} \mathbf{V}^\top, \text{with } \mathbf{U}^{\top} \mathbf{U} = \mathbf{I}, \mathbf{V}^{\top} \mathbf{V} = \mathbf{I} \quad (\text{SVD}) \\
    & \mathbf{F}=\mathbf{R}\mathbf{S}, \text{with } \mathbf{S}=\mathbf{S}^{\top}, \mathbf{R} \mathbf{R}^{\top} = \mathbf{I} \quad (\text{Polar decomposition}) \\
\end{aligned}
\end{equation}

The \textit{spatial gradient of velocity} or \textit{velocity gradient tensor} $\mathbf{L}$ is defined as:

\begin{equation}
    \mathbf{L}(\mathbf{x}, t) := \frac{\partial \mathbf{v}}{\partial \mathbf{x}}
\end{equation}

The \textit{material time derivative of the deformation gradient} $\dot{\mathbf{F}}$ is given as \footnote{The commuting property of the material time derivative of Lagrangian fields with material gradient, which doesn't hold for Eulerian fields, has been exploited to derive this formula relating the deformation gradient tensor \(\mathbf{F}\), its time derivative \(\dot{\mathbf{F}}\), the velocity gradient tensor \(\mathbf{L}\), and the spatial and material coordinates.}:

\begin{equation}
    \dot{\mathbf{F}} = \frac{\partial}{\partial t} \left( \frac{\partial \phi(\mathbf{X}, t)}{\partial \mathbf{X}} \right) = \frac{\partial \mathbf{v}}{\partial \mathbf{X}} = \frac{\partial \mathbf{v}}{\partial \mathbf{x}} \cdot \frac{\partial \mathbf{x}}{\partial \mathbf{X}} = \mathbf{L} \cdot \mathbf{F}
\end{equation}

The \textit{material time derivative} $D /D t$ has two formulations. Using density $\rho(\mathbf{X}, t)$ as an example, the Lagrangian description prescribes its material time derivative as:

\begin{equation}
\frac{D \rho(\mathbf{X}, t)}{D t} \equiv \dot{\rho} = \frac{\partial \rho(\mathbf{X}, t)}{\partial t},
\end{equation}

\noindent the Eulerian description calculates it as:

\begin{equation}
\begin{aligned}
    \frac{D \rho(\mathbf{x}, t)}{D t} \equiv \dot{\rho} &:= \lim_{\Delta t \to 0} \frac{\rho(\phi(\mathbf{X}, t + \Delta t), t + \Delta t) - \rho(\phi(\mathbf{X}, t), t)}{\Delta t} \\
    &= \frac{\partial \rho(\mathbf{x}, t)}{\partial t} + \nabla \rho(\mathbf{x}, t) \cdot \mathbf{v}(\mathbf{x}, t)
\end{aligned}
\end{equation}

\noindent as a result of the chain rule of derivative, referencing Eq.\ref{eq:Lagrangian_Eulerian_coordinates}. 

Commonly used strain measures in continuum mechanics are \textit{the right Cauchy-Green deformation tensor}, \textit{the Green strain tensor}, and \textit{the rate of deformation tensor}. Their definitions can be found in \cite{DeVaucorbeil2020}. In accordance with the strain measures, multiple stress measures are used, including \textit{Cauchy stress}, \textit{Kirchhoff stress} (the weighted Cauchy stress), \textit{first Piola-Kirchhoff stress} (1st PK), \textit{second Piola-Kirchhoff stress} (2nd PK). The stresses inside a solid body or fluid are described by a tensor field, stress tensor and strain tensor are both second-order tensor (matrix) fields, e.g. if they live in a 3D body, there are 9 components (each of the three faces of a cube-shaped infinitesimal segment has 3 stress components) which can be conveniently represented as a $3 \times 3$ array. The relation between the four types of stresses are summarised in Table.\ref{table:stress_relations}.

\begin{table}[H]
\centering
\small
\renewcommand{\arraystretch}{1.5}
\begin{tabular}{|c|c|c|c|c|}
\hline
 & $\bm{\sigma}$ & $\bm{\tau}$ & $\mathbf{P}$ & $\mathbf{S}$ \\
\hline
Cauchy stress $\bm{\sigma}$ & - & $\bm{\tau} J^{-1}$ & $J^{-1} \mathbf{P} \mathbf{F}^{\top}$ & $J^{-1} \mathbf{F} \mathbf{S} \mathbf{F}^{\top}$ \\
\hline
Kirchhoff stress $\bm{\tau}$ & $J \bm{\sigma}$ & - & $\mathbf{P} \mathbf{F}^{\top}$ & $\mathbf{F} \mathbf{S} \mathbf{F}^{\top}$ \\
\hline
1\textsuperscript{st} PK $\mathbf{P}$ & $J \bm{\sigma} \mathbf{F}^{-\top}$ & $\bm{\tau} \mathbf{F}^{-\top}$ & - & $\mathbf{F} \mathbf{S}$ \\
\hline
2\textsuperscript{nd} PK $\mathbf{S}$ & $J \mathbf{F}^{-1} \bm{\sigma} \mathbf{F}^{-\top}$ & $\mathbf{F}^{-1} \bm{\tau} \mathbf{F}^{-\top}$ & $\mathbf{F}^{-1} \mathbf{P}$ & - \\
\hline
\end{tabular}
\caption{Relation between different stress measures (table from \cite{DeVaucorbeil2020}).}
\label{table:stress_relations}
\end{table}

Three types of \textit{conservation equations} are commonly used in continuum mechanics: conservation of mass, momentum, and energy. \textit{Equation of mass conservation}, or the \textit{continuity equation} can be expressed as:

\begin{equation} \label{eq:conservation_of_mass}
\frac{D \rho}{D t} + \rho \nabla \cdot \mathbf{v} = 0 \quad \text{(Continuity equation)}
\end{equation}

\noindent where $\rho(\mathbf{X},t)$ is the mass density, $\mathbf{v}(\mathbf{X},t)$ is the velocity. Some literature also write the continuity equation as 

\begin{equation} \label{eq:conservation_of_mass2} \tag{\ref{eq:conservation_of_mass}b}
    \frac{\partial \rho}{\partial t} + \nabla \cdot (\rho \mathbf{v}) = 0
\end{equation}

\noindent if we expand the divergence operator $\nabla \cdot (\rho \mathbf{v}) = (\nabla \rho) \cdot \mathbf{v} + \rho (\nabla \cdot \mathbf{v})$ and note the material derivative $\frac{D \rho}{D t} = \frac{\partial \rho}{\partial t} + (\nabla \rho) \cdot \mathbf{v}$, then equivalence between Eq.\ref{eq:conservation_of_mass2} and Eq.\ref{eq:conservation_of_mass} is observed. 

If the material is incompressible, i.e. $\frac{D \rho}{D t}=0$, which simplifies Eq.\ref{eq:conservation_of_mass} to $\nabla \cdot \mathbf{v} = 0$. For Lagrangian description, a simpler algebraic equation of mass conservation is \cite{DeVaucorbeil2020}:

\begin{equation} \label{eq:conservation_of_mass_Lagrangian} \tag{\ref{eq:conservation_of_mass}b}
    \rho J = \rho_0
\end{equation}

Conservation of (linear) momentum \footnote{Conservation of angular momentum states that the Cauchy stress must be a symmetric tensor.} states that, change of linear momentum over time is induced by net force (the sum of all volume and surface forces) acting on the body:

\begin{equation} \label{eq:conservation_of_linear_momentum}
\rho \frac{D \mathbf{v}}{D t} = \nabla \cdot \bm{\sigma} + \mathbf{f}^{ext} \quad \text{(Linear momentum equation)}
\end{equation}

\noindent where $\bm{\sigma}(\mathbf{X},t)$ is the symmetric \footnote{Symmetry is required by conservation of \textit{angular} momentum.} Cauchy stress tensor. $\mathbf{f}^{ext}$ is the external body force, for example, the gravitational force $\mathbf{f}^{ext}=\rho \mathbf{g}$ with $\mathbf{g}$ being the gravitational acceleration.

Conservation of energy, e.g. for a thermo-mechanical system, states that, the total energy change in the body equals the work done by external forces and work provided by heat flux $\mathbf{q}$ and other energy sources. As we are not concerned about the thermal behaviour in this context, the energy equation is not presented.

\paragraph{Constitutive models} While the former equations are universal to both solids and fluids, it is the constitutive model, e.g. Hooke’s law for linear elastic materials, which relates kinetic quantities (e.g. stresses) to kinematic quantities (e.g. strains), that differentiates them. In engineering, a constitutive model expresses the relation between stress and strain (or between force and deformation). The constitutive model for \textit{linear elastic isotropic materials}, for example, writes \cite{DeVaucorbeil2020}: 

\begin{equation} \label{eq:stress_strain_linear_elastic_material}
    \boldsymbol\sigma = (\lambda \text{tr} \boldsymbol\epsilon) \mathbf{I} + 2\mu \boldsymbol\epsilon
\end{equation}

\noindent where $\boldsymbol\sigma$ and $\boldsymbol\epsilon$ are stress and strain, respectively. $\lambda$ and $\mu$ are the first and second Lamé's constant, respectively.

In continuum mechanics, it is common for the constitutive model to relate the \textit{deformation gradient} $\mathbf{F}$ and the \textit{first Piola-Kirchoff stress} (1st PK) $\mathbf{P}$ as they are more naturally expressed in material space \cite{Jiang2016}. Hyper-elastic materials, for example, have \cite{Jiang2016}

\begin{equation} \label{eq:1stPK_stress_deformation_gradient}
    \mathbf{P} = \frac{\partial \Psi(\mathbf{F})}{\partial \mathbf{F}}
\end{equation}

\noindent where $\Psi$ is the scalar-valued elastic energy density function. $\mathbf{P}$ is a matrix with same dimension as $\mathbf{F}$. $\mathbf{P}$ can be related to the \textit{Cauchy stress} $\boldsymbol{\sigma}$, which is more commonly used in engineering literature, as follows (see Table.\ref{table:stress_relations}):

\begin{equation} \label{eq:Cauchy_stress_deformation_gradient}
\boldsymbol{\sigma} = \frac{1}{J} \mathbf{P} \mathbf{F}^\top = \frac{1}{\det(\mathbf{F})} \frac{\partial \Psi(\mathbf{F})}{\partial \mathbf{F}} \mathbf{F}^\top
\end{equation}

\noindent in the RHS, the stress tensor $\mathbf{P}=\frac{\partial \Psi(\mathbf{F})}{\partial \mathbf{F}}$ is a second-order tensor (matrix), $\mathbf{F}^\top$ is also a secind-order tensor, their product also results in again a second-order tensor. The energy density function $\Psi(\mathbf{F})$ penalizes non-rigid $\mathbf{F}$, $\Psi(\mathbf{F})$=0 if $\mathbf{F}$ is a rotation.

The energy density function of the \textit{Neo-Hookean model} \footnote{The Neo-Hookean model is a non-linear hyper-elastic model for predicting large deformations of elastic materials \cite{Jiang2016}.} as an example, is \cite{Jiang2016}:

\begin{equation} \label{eq:Neo_Hookean_energy_density_func}
\Psi(\mathbf{F}) = \frac{\mu}{2} \left( \text{tr}(\mathbf{F}^\top \mathbf{F}) - d \right) - \mu \log(J) + \frac{\lambda}{2} \log^2(J)
\end{equation}

\noindent where $d$ is the dimension, $\mu$ is the first Lamé parameter (i.e. the shear modulus), and $\lambda$ the second Lamé parameter. They can be related to \textit{Young's modulus} $E$ and \textit{Poisson's ratio} $\nu$ via

\begin{equation} \label{eq:shear_modulus_Lame_constant}
\mu = \frac{E}{2(1 + \nu)}, \quad \lambda = \frac{E \nu}{(1 + \nu)(1 - 2 \nu)}
\end{equation}

\noindent we can then use the relation $\mathbf{P} = \frac{\partial \Psi(\mathbf{F})}{\partial \mathbf{F}}$ (Eq.\ref{eq:1stPK_stress_deformation_gradient}) to derive the 1st PK stress, as well as the Cauchy stress (via Eq.\ref{eq:Cauchy_stress_deformation_gradient}):

\begin{equation} \label{eq:1stPKStress_CauchyStress_F}
    \mathbf{P} = \mu (\mathbf{F} - \mathbf{F}^{-\top}) + \lambda \log(J) \mathbf{F}^{-\top}, \quad \boldsymbol{\sigma} = \frac{1}{J} [\mu (\mathbf{F}\mathbf{F}^\top - \mathbf{I}) + \lambda \log(J) \mathbf{I}]
\end{equation}

Common solid constitutive models for modelling different types of materials (e.g. elastic, hyper-elastic, plastic, visco-elastic, elasto-plastic) can be found in monographs on mechanics such as \cite{Timoshenko1951,Bennett2006} as well as MPM literature such as \cite{Jiang2015thesis,DeVaucorbeil2020}. 

\paragraph{Particle-in-cell (PIC) transfer} Here we follow pp.42-43 in \cite{Jiang2015thesis}. At any iteration (ignoring the iteration number superscript $n$), we transfer mass and momentum from particles to grid nodes: 

\begin{equation} \label{eq:PIC}
    m_i = \sum_{p=1}^{M} w_{ip} m_p, \quad m_i \mathbf{v}_i = \sum_{p=1}^{M} w_{ip} m_p \mathbf{v}_p
\end{equation}

\noindent where $w_{ip}=K(\mathbf{x}_p - \mathbf{x}_i)$ are the interpolation weights. We can then calculate the grid nodal velocity $\mathbf{v}_i$ and update it using the forces acting on the particle (Newton's second law). The updated nodal velocity is then interpolated back to the particle to obtain particle velocity:

\begin{equation} \label{eq:particle_velocity}
    \mathbf{v}_p = \sum_{i=1}^N w_{ip} \mathbf{v}_i
\end{equation}

\noindent where $N$ is the total number of grid nodes. The issue with PIC is loss of angular momentum, which leads to rotational motion damping.

\paragraph{Affine particle-in-cell (APIC) transfer} Here we mainly follow pp.41-42 in \cite{Jiang2016}. At any iteration, the transfer of mass and momentum from particles to grid, derived by preserving affine motion during the transfers, is:

\begin{equation} \label{eq:APIC}
m_i = \sum_{p=1}^{M} w_{ip} m_p, \quad m_i \mathbf{v}_i = \sum_{p=1}^{M} w_{ip} m_p [\mathbf{v}_p + \mathbf{B}_p (\mathbf{D}_p)^{-1} (\mathbf{x}_i - \mathbf{x}_p)]
\end{equation}

\noindent where \(w_{ip}\) is the interpolation weight for particle \(p\) and grid node \(i\). The particle affine tensor $\mathbf{B}_p$, stored at each particle, is a $d \times d$ matrix which represents an affine transformation that approximates the local, linear velocity gradients (i.e., how velocity changes in space around the particle). $\mathbf{D}_p$ is calculated as

\begin{equation} \label{eq:APIC_Dp}
    \mathbf{D}_p = \sum_{i=1}^{N} w_{ip} (\mathbf{x}_i - \mathbf{x}_p) (\mathbf{x}_i - \mathbf{x}_p)^\top
\end{equation}

\noindent which is a $d \times d$ matrix capturing the spatial distribution of the grid nodes around the particle. For quadratic interpolation stencils, we have $\mathbf{D}_p = \frac{1}{4} h^2 \mathbf{I}$; for cubic interpolation stencils, $\mathbf{D}_p = \frac{1}{3} h^2 \mathbf{I}$, with $h$ being the grid spacing. Therefore, $(\mathbf{D}_p)^{-1}$ becomes a scaling factor in Eq.\ref{eq:APIC}.

Inversely, we have the transfer from grid back to particles:

\begin{equation} \tag{cc.Eq.\ref{eq:particle_velocity}}
    \mathbf{v}_p = \sum_{i=1}^{N} w_{ip} \mathbf{v}_i
\end{equation}

\begin{equation} \label{eq:APIC_Bp}
    \mathbf{B}_p = \sum_{i=1}^{N} w_{ip} \mathbf{v}_i (\mathbf{x}_i - \mathbf{x}_p)^\top
\end{equation}

Note that, the motion of grid nodes are virtual - we never compute a new grid, we only explicitly store the grid velocities \cite{Jiang2016}.

\paragraph{Forces} Here we mainly follow pp.43-44 in \cite{Jiang2016}. Forces are calculated on grid nodes. At iteration $n$, the internal force $\mathbf{f}_i^{int,n}$ acting on grid node $i$ can be derived from the weak form of momentum equation: 

\begin{equation} \label{eq:MPM_grid_forces_Cauchy_stress}
\mathbf{f}_i^{int,n} = \mathbf{f}_i^{int}(\mathbf{x}_i) = - \sum_{p=1}^{M} V_p^n \boldsymbol{\sigma}_p^n \nabla w_{ip}^n
\end{equation}

\noindent where $V_p^n$, $\boldsymbol{\sigma}_p^n$ are the volume of and Cauchy stress acting on particle $p$, respectively.

Forces can also be derived via total potential energy \cite{Jiang2016}, i.e. aggregating all stress (Eq.\ref{eq:Cauchy_stress_deformation_gradient}) contributions:

\begin{equation} \label{eq:MPM_grid_forces_energy_density}
\mathbf{f}_i^{int,n} = \mathbf{f}_i^{int}(\mathbf{x}_i) = - \sum_{p=1}^{M} V_p^0 \left[\frac{\partial \Psi(\mathbf{F}_p^n)}{\partial \mathbf{F}} \right] (\mathbf{F}_p^n)^\top \nabla w_{ip}^n
\end{equation}

\noindent where $V_p^0$ is the volume originally occupied by the particle $p$, it can be related to $V_p^n$ via $V_p^n=\det(\mathbf{F}_p^n) V_p^0$, which makes Eq.\ref{eq:MPM_grid_forces_energy_density} equivalent to Eq.\ref{eq:MPM_grid_forces_Cauchy_stress} with reference to Eq.\ref{eq:Cauchy_stress_deformation_gradient}. $\Psi$ is the energy density function. The summation is over all particles $p$ that contribute to the force at grid node $i$. On the RHS, the product $\left[\frac{\partial \Psi(\mathbf{F}_p^n)}{\partial \mathbf{F}} \right] (\mathbf{F}_p^n)^\top$ results in a second-order tensor (stress tensor); $\nabla w_{ip}^n$ is a vector.

\paragraph{Update of deformation gradient $\mathbf{F}$ } Here we mainly follow pp.42-43 in \cite{Jiang2016}. At any iteration $t=n$, we can update the deformation gradient $\mathbf{F}$ for each particle as follows \cite{Jiang2016}: 

\begin{equation} \label{eq:MPM_deformaiton_gradients_update}
\mathbf{F}_p^{n+1} = \left( \mathbf{I} + \Delta t \cdot \sum_{i=1}^{N} \mathbf{v}_i^{n+1} (\nabla w_{ip}^n)^\top \right) \mathbf{F}_p^n
\end{equation}

\section{Interpolation kernels} \label{app:interpolation_kernels}

Here we follow pp.33 in \cite{Jiang2016} and present 3 commonly used interpolation kernels with $r=(x_p-x_i)/h$. Some other interpolation functions can be found in e.g. \cite{DeVaucorbeil2020}.

1. Linear kernel

\begin{equation}
    K(r) = 
    \begin{cases} 
    1 - |r| & 0 \leq |r| < 1, \\ 
    0 & 1 \leq |r|.
    \end{cases}
\end{equation}

2. Quadratic kernel

\begin{equation}
    K(r) = 
    \begin{cases} 
    \frac{3}{4} - |r|^2 & 0 \leq |r| < \frac{1}{2}, \\ 
    \frac{1}{2} \left(\frac{3}{2} - |r|\right)^2 & \frac{1}{2} \leq |r| < \frac{3}{2}, \\ 
    0 & \frac{3}{2} \leq |r|.
    \end{cases}
\end{equation}  

3. Cubic kernel

\begin{equation}
    K(r) = 
    \begin{cases} 
    \frac{1}{2}|r|^3 - |r|^2 + \frac{2}{3} & 0 \leq |r| < 1, \\ 
    \frac{1}{6}(2 - |r|)^3 & 1 \leq |r| < 2, \\ 
    0 & 2 \leq |r|.
    \end{cases}
\end{equation}

\section{Alternative MPM-ParVI implementation}

\begin{algorithm}[H]
\fontsize{8}{8}
\caption{MPM-based sampling (APIC transfer, part 1)}
\label{algo:MPM_sampling2}
\begin{itemize}
\item{\textbf{Inputs}: a differentiable target density $p(\mathbf{x})$ with magnitude amplification constant $\alpha$, total number of particles $M$, particle (density) dimension $d$, initial proposal distribution $p^0(\mathbf{x})$, total number of grid nodes $N$, total number of iterations $T$, step size $\Delta t$. An energy density function $\Psi$ specifying the constitutive model, and an interpolation function $K$.} 
\item{\textbf{Outputs}: Particles located at positions $\{\mathbf{x}_p \in \mathbb{R}^d \}_{p=1}^{M}$ whose empirical distribution approximates the target density $p(\mathbf{x})$.}
\end{itemize}
\vskip 0.05in
1. \textbf{\textit{Initialise}} particles and grid. [$\mathcal{O}(max(M,N)d)$]
    \begin{addmargin}[1em]{0em}% 
    Initialise $M$ particles from $p^0(\mathbf{x})$, each with mass $m_p$, volume $V_p^0$, initial position $\mathbf{x}_p$, initial velocity $\mathbf{v}_p$, affine matrix $\mathbf{B}_p$, and material-related parameters. Some default values can be set to be 0 \cite{Jiang2015thesis}. [$\mathcal{O}(Md)$] \\ 
    Initialise a grid with $N$ nodes. Grid node locations $\mathbf{x}_i, i=1,2,...,N$ are on a undeformed regular grid and are time-invariant. Initialise grid data, i.e. nodal mass $m_i$, velocity $\mathbf{v}_i$, to default values of 0. [$\mathcal{O}(Nd)$]
    \end{addmargin}

2. \textbf{\textit{Update}} particle positions. \\
For each iteration $t=1,2,...,T$, repeat until optimal configuration:
    \begin{addmargin}[1em]{0em}% 
    (1) Compute the interpolation weights $w_{ip}^n$ and weight gradients $\nabla w_{ip}^n$ for each particle $p$:
    \[
    w_{ip}^n=K(\mathbf{x}_i - \mathbf{x}_p^n), \quad \nabla w_{ip}^n=\nabla K(\mathbf{x}_i - \mathbf{x}_p^n)
    \]
    These location-dependent weights and weight gradients are computed once and stored on the particles in each iteration. \\
    (2) Transfer \textbf{particle} quantities to the \textbf{grid} (\textbf{\textit{P2G.}}). Compute the \textbf{grid} mass and momentum using APIC (Eq.\ref{eq:APIC}): [$\mathcal{O}(Md^3+MNd^2)$]
    \[
    m_i^n = \sum_{p=1}^{M} w_{ip}^n m_p, \quad m_i^n \mathbf{v}_i^n = \sum_{p=1}^{M} w_{ip}^n m_p [\mathbf{v}_p^n + \mathbf{B}_p^n (\mathbf{D}_p^n)^{-1} (\mathbf{x}_i - \mathbf{x}_p^n)]
    \]
    where $\mathbf{D}_p^n= \sum_{i=1}^{N} w_{ip}^n (\mathbf{x}_i - \mathbf{x}_p^n) (\mathbf{x}_i - \mathbf{x}_p^n)^\top$ (Eq.\ref{eq:APIC_Dp}). \\ 
    (3) Compute \textbf{grid} velocities (Eq.\ref{eq:grid_velocity}): [$\mathcal{O}(Nd)$]
    \[
   \mathbf{v}_i^n = \frac{m_i^n \mathbf{v}_i^n}{m_i^n} 
    \]
    (4) Compute explicit \textbf{grid} forces $\mathbf{f}_i^{int,n}$ (Eq.\ref{eq:MPM_grid_forces_Cauchy_stress} or Eq.\ref{eq:MPM_grid_forces_energy_density}): [$\mathcal{O}(Md^3+MNd^2)$] \\
    \[
    \mathbf{f}_i^{int,n} = - \sum_{p=1}^{M} V_p^n \boldsymbol{\sigma}_p^n \nabla w_{ip}^n, \quad \text{or} \quad
    \mathbf{f}_i^{int,n} = - \sum_{p=1}^{M} V_p^0 [\frac{\partial \Psi(\mathbf{F}_p^n)}{\partial \mathbf{F}} ] (\mathbf{F}_p^n)^\top \nabla w_{ip}^n
    \]
    where $\boldsymbol{\sigma}_p^n = \frac{1}{\det(\mathbf{F})} \frac{\partial \Psi(\mathbf{F}_p^n)}{\partial \mathbf{F}} (\mathbf{F}_p^n)^\top$ (Eq.\ref{eq:Cauchy_stress_deformation_gradient}). \\
    \newpage
    \end{addmargin}
\end{algorithm}

\begin{algorithm}[H] \ContinuedFloat
\fontsize{8}{8}
\caption{MPM-based sampling (APIC transfer, part 2)}
\begin{addmargin}[1em]{0em}%
    (5) Update \textbf{grid} velocities $\mathbf{v}_i^{n}$ using a explicit Euler time integration scheme (Eq.\ref{eq:Eulerian_time_integration}): [$\mathcal{O}(Nd)$]
    \[
    \mathbf{v}_i^{n+1} = \mathbf{v}_i^n + \Delta t \cdot (\mathbf{f}_i^{int,n} + \mathbf{f}_i^{ext}) / m_i
    \]
    where the time-invariant, external body force $\mathbf{f}_i^{ext}=\alpha \nabla_{\mathbf{x}} \log p(\mathbf{x}_i)$. \\
    (6) Transfer \textbf{grid} property states to \textbf{particles} (\textbf{\textit{G2P}}). Compute the new \textbf{particle} velocities $\mathbf{v}_p^{n+1}$ (Eq.\ref{eq:particle_velocity}) and positions $\mathbf{x}_p^{n+1}$; update \textbf{particle} affine matrices $\mathbf{B}_p^{n+1}$ (Eq.\ref{eq:APIC_Bp}) and deformation gradients $\mathbf{F}_p^{n}$ (Eq.\ref{eq:MPM_deformaiton_gradients_update}): [$\mathcal{O}(Md^3+MNd^2)$] \\
    \[
    \mathbf{v}_p^{n+1} = \sum_{i=1}^{N} w_{ip}^{n} \mathbf{v}_i^{n+1}, \quad \mathbf{x}_p^{n+1} = \mathbf{x}_p^{n} + \Delta t \cdot \mathbf{v}_p^{n+1}
    \]
    \[
    \mathbf{B}_p^{n+1} = \sum_{i=1}^{N} w_{ip}^{n} \mathbf{v}_i^{n+1} (\mathbf{x}_i - \mathbf{x}_p^{n})^\top, \quad \mathbf{F}_p^{n+1} = \left( \mathbf{I} + \Delta t \cdot \sum_{i=1}^{N} \mathbf{v}_i^{n+1} (\nabla w_{ip}^n)^\top \right) \mathbf{F}_p^n
    \]
    (7) Re-set grid data, i.e. nodal mass $m_i$, velocity $\mathbf{v}_i$, to the default values of 0.
    \end{addmargin}

3. \textbf{\textit{Return}} the intermediate or final particle positions $\{\mathbf{x}_p^{T}\}_{p=1}^{M}$ and their histogram and/or \textit{KDE} estimate for each dimension. \\
\end{algorithm}

\end{document}